\documentclass[lettersize,journal]{IEEEtran}
\usepackage{amsmath,amsfonts}
\usepackage{array}
\usepackage[caption=false,font=normalsize,labelfont=sf,textfont=sf]{subfig}
\usepackage{textcomp}
\usepackage{stfloats}
\usepackage{url}
\usepackage{verbatim}
\usepackage{graphicx}
\def\BibTeX{{\rm B\kern-.05em{\sc i\kern-.025em b}\kern-.08em
    T\kern-.1667em\lower.7ex\hbox{E}\kern-.125emX}}
\usepackage{balance}
\usepackage{multirow}
\usepackage{booktabs} 
\usepackage{hyperref}
\usepackage{bm}
\usepackage{makecell}
\usepackage{manyfoot}%
\usepackage{threeparttable}
\usepackage{tabu}
\usepackage{tabularx}
\usepackage{algorithm}%
\usepackage{algorithmicx}
\usepackage{algpseudocode}%
\usepackage{listings}%
\usepackage{xcolor}%
\usepackage[normalem]{ulem}

\usepackage{amsthm} 

\theoremstyle{plain}
\newtheorem{theorem}{Theorem}[section] 
\newtheorem{lemma}[theorem]{Lemma}

\theoremstyle{definition}
\newtheorem{assumption}{Assumption}    
\numberwithin{assumption}{section}

\theoremstyle{remark}
\newtheorem{remark}{Remark}[section]

\begin{document}
\title{Learning Evolution via Optimization Knowledge Adaptation}
\author{Chao Wang,~\IEEEmembership{Member,~IEEE}, Lingling Li,~\IEEEmembership{Senior Member,~IEEE}, Licheng Jiao,~\IEEEmembership{Fellow,~IEEE}, Jiaxuan Zhao,~\IEEEmembership{Student Member,~IEEE}, Fang Liu,~\IEEEmembership{Senior Member,~IEEE}, Shuyuan Yang,~\IEEEmembership{Senior Member,~IEEE}

\thanks{This work was supported in part by the Joint Funds of the National Natural Science Foundation of China (U22B2054), the National Natural Science Foundation of China (62076192, 62276199, 62431020 and 62276201), the 111 Project, the Program for Cheung Kong Scholars and Innovative Research Team in University (IRT 15R53), the Science and Technology Innovation Project from the Chinese Ministry of Education. the National Key Laboratory of Human-Machine Hybrid Augmented Intelligence, Xi'an Jiaotong University (HMHAI-202404and HMHAI-202405). (\textit{Corresponding author: Lingling Li}.)}
\thanks{The authors are with the Key Laboratory of Intelligent Perception and Image Understanding of Ministry of Education, International Research Center for Intelligent Perception and Computation, Xidian University, Xi’an 710071, China (e-mail: llli@xidian.edu.cn).}}

\markboth{Journal of \LaTeX\ Class Files,~Vol.~1, No.~1, January~2025}%
{Learning Evolution via Optimization Knowledge Adaptation}

\maketitle

\begin{abstract}
\textcolor{black}{The iterative search process of evolutionary algorithms (EAs) encapsulates optimization knowledge within historical populations and fitness evaluations. Effective utilization of this knowledge is crucial for facilitating knowledge transfer and online adaptation. However, current research typically addresses these goals in isolation and faces distinct limitations: evolutionary sequential transfer optimization often suffers from incomplete utilization of prior knowledge, while adaptive strategies, utilizing real-time knowledge, are limited to tailoring specific evolutionary operators. To simultaneously achieve these two capabilities, we introduce the Optimization Knowledge Adaptation Evolutionary Model (OKAEM), a unified learnable evolutionary framework capable of adaptively updating parameters based on available optimization knowledge. By parameterizing evolutionary operators via attention mechanisms, OKAEM enables learnable update rules that facilitate the utilization of optimization knowledge via two phases: pre-training to integrate extensive prior knowledge for efficient transfer, and adaptive optimization to dynamically update parameters based on real-time knowledge. Experimental results confirm that OKAEM significantly outperforms state-of-the-art sequential transfer methods across 12 transfer scenarios via pre-training, and surpasses advanced learnable EAs solely through its self-tuning mechanism in prior-free settings. Beyond demonstrating practical utility in prompt tuning for vision-language models, ablation studies validate the necessity of the learnable components, while visualization analyses reveal the model's capacity to autonomously discover interpretable evolutionary principles.} \footnote{The code can be accessed at \url{https://gitee.com/Anonymity_Paper/code-of-okaem}.}
\end{abstract}


\section{Introduction}
\IEEEPARstart{I}{nspired} by biological evolution, evolutionary algorithms (EAs) continuously update population systems through crossover, mutation, and selection to explore complex fitness landscapes \cite{10}. Prominent examples of EAs include the genetic algorithm (GA) \cite{2}, the evolution strategy (ES) \cite{3}, and the genetic programming (GP) \cite{4}. These methods rely solely on the fitness values of individuals to drive the evolutionary process without requiring gradient information. Advances in computational techniques \cite{5} have allowed EAs to provide diverse solutions for highly complex optimization tasks such as neuroevolution \cite{6,yan2024populating}, robotic control \cite{7,slade2022personalizing}, industrial design \cite{8}, and scientific discoveries \cite{9,doi:10.1073/pnas.2318641121}. As the scale and complexity of optimization tasks increase \cite{10,1}, \textcolor{black}{facilitating knowledge transfer to leverage prior experience and online adaptation to robustly adjust to complex fitness landscapes is crucial for enhancing the optimization capabilities of EAs.}

\textcolor{black}{Regarding knowledge transfer, evolutionary sequential transfer optimization (ESTO)} \cite{11,12,1324694,8733017,14,15,9385398} aims to accelerate evolutionary optimization for challenging target tasks by leveraging prior experience accumulated from source tasks. Despite its success, current mainstream ESTO methods primarily focus on knowledge derived from highly related individuals \cite{21}, thereby neglecting valuable holistic information from other candidates within the population. \textcolor{black}{The incomplete utilization of prior knowledge} hampers the full exploitation of underlying evolutionary behaviors on source tasks. Furthermore, \textcolor{black}{online adaptation} has long been a key challenge in the field of EAs \cite{2,Turing2009,16,17}. \textcolor{black}{Traditional adaptive strategies primarily rely on real-time evolutionary dynamics, such as the current population distribution and fitness evaluations, to adjust the parameters or optimization strategies.} However, the absence of a unified framework for heuristic evolutionary operators results in adaptation strategies specific to particular EAs. For example, the covariance matrix adaptation used in CMA-ES \cite{18} may not be effective in GA contexts. \textcolor{black}{While recent learnable EAs (LEAs) \cite{19,20,NEURIPS2019_ec04e8eb,hong2023pre,li2023b2opt} aim to enhance adaptability and generalizability by training parameterized operators, most existing models fix their parameters after meta-training. The lack of online adaptation prevents them from dynamically adjusting to real-time evolutionary dynamics, limiting their effectiveness in unknown landscapes.}

\textcolor{black}{Although knowledge transfer and online adaptation are typically studied in isolation, addressing their respective limitations relies on a common foundation: the effective utilization of optimization knowledge. Optimization knowledge comprises historical populations and fitness evaluations, encapsulating critical search trajectories and landscape characteristics. In terms of utilization, ESTO leverages prior knowledge from source tasks to accelerate convergence, whereas adaptive EAs rely on real-time knowledge to dynamically adjust search strategies. Consequently, establishing a unified paradigm that seamlessly integrates these two knowledge sources is essential to develop an optimizer capable of simultaneously exploiting prior experience for efficient transfer and utilizing real-time feedback for robust self-adaptation.}

\begin{figure*}[htbp]
\centering
\includegraphics[width=0.9\textwidth]{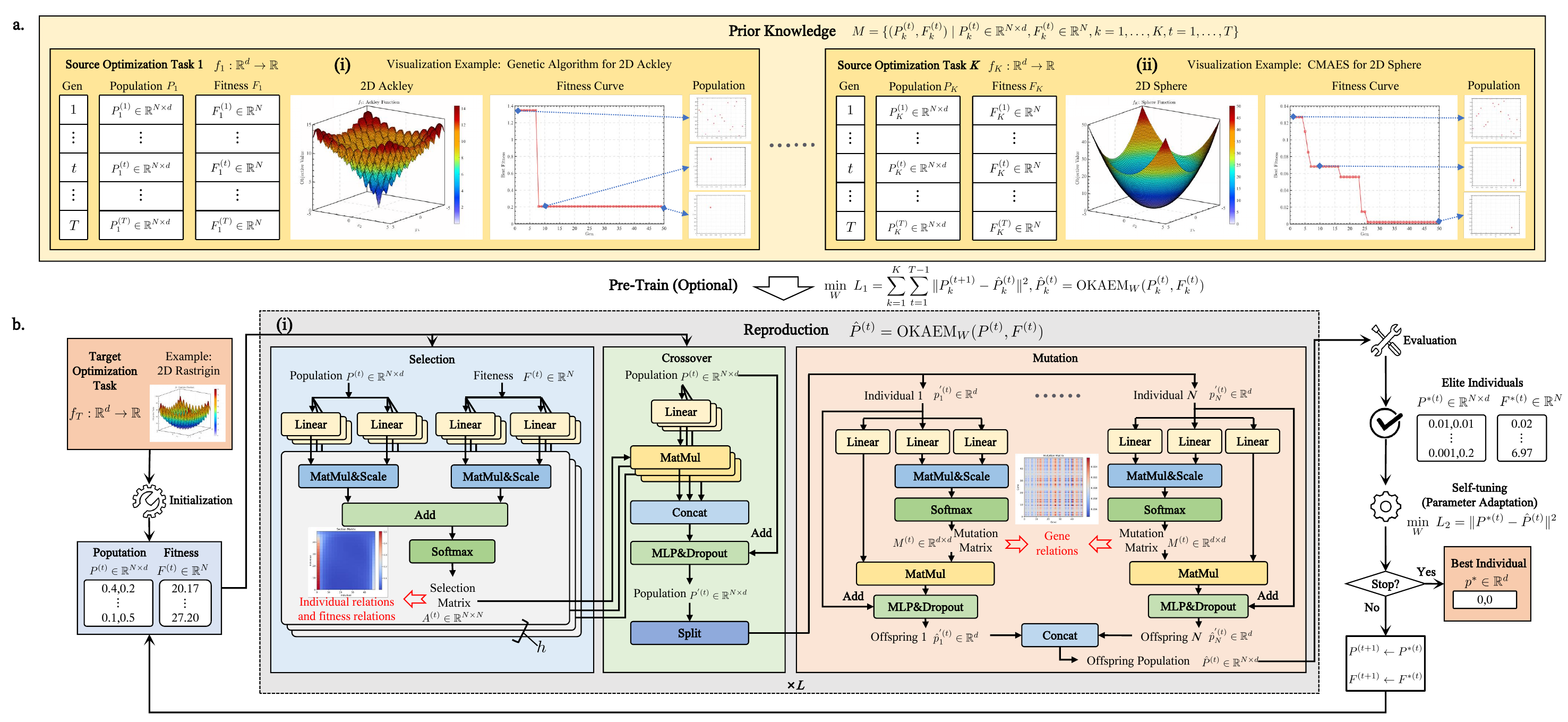}
\caption{Schematic flow of OKAEM. a. Pre-training on prior knowledge (optional): (i) example: prior knowledge accumulated using a genetic algorithm on the 2D-Ackley problem; (ii) example: prior knowledge accumulated using CMAES on the 2D-Sphere problem. b. Adaptive optimization to solve target tasks. This phase includes initialization, reproduction, evaluation, elitism, and self-tuning (parameter adaptation): (i) selection, crossover, and mutation modules used in reproduction.}\label{fig1}
\end{figure*}

\textcolor{black}{To simultaneously achieve knowledge transfer and online adaptation}, we introduce the Optimization Knowledge Adaptation Evolutionary Model (OKAEM), \textcolor{black}{a unified learnable evolutionary framework capable of adaptively updating parameters based on available optimization knowledge.} \textcolor{black}{Unlike existing methods that separate transfer and adaptation, OKAEM seamlessly integrates extensive prior knowledge with real-time knowledge.}
By leveraging attention mechanisms, OKAEM explicitly models the relationships among individuals, fitness landscapes, and genetic components, guiding \textcolor{black}{the parameterization of} selection, crossover, and mutation. \textcolor{black}{These parameterized evolutionary operators enable learnable and highly parallelizable update rules, facilitating the utilization of optimization knowledge via two phases: pre-training and adaptive optimization (Fig. \ref{fig1}).} During the pre-training phase, OKAEM learns population evolution behaviors on source tasks by predicting the next generation's population, thereby leveraging prior knowledge to enhance performance. In the adaptive optimization phase, OKAEM continuously generates offspring and dynamically updates its parameters based on \textcolor{black}{real-time knowledge} by minimizing the distance between generated populations and elite individuals. \textcolor{black}{The novel combination of pre-training and self-tuning mechanisms effectively facilitates knowledge transfer and online adaptation, addressing the limitations of incomplete prior knowledge utilization and customized adaptation strategies.} Notably, even without prior knowledge, users can still execute the adaptive optimization phase to solve target tasks effectively.

\textcolor{black}{Experimental evaluations focus on two core capabilities: knowledge transfer and online adaptation. In terms of knowledge transfer, OKAEM utilizes pre-training to significantly outperform classic and state-of-the-art ESTO methods across 12 scenarios. Regarding online adaptation, OKAEM surpasses advanced LEAs solely through its self-tuning mechanism, even in the absence of prior knowledge.} Leveraging parallel computing, OKAEM requires only a few GPU seconds. In a practical case study on prompt tuning for vision-language models, OKAEM surpasses state-of-the-art black-box baselines. Moreover, we find that OKAEM improves its performance with knowledge accumulation and explicitly learns the principles of natural selection and genetic recombination.

The rest of this paper is organized as follows. Related \textcolor{black}{work on ESTO} and LEAs \textcolor{black}{is} introduced in Section II. Section III offers a detailed exposition of the proposed OKAEM. In Section IV, we validate the effectiveness of OKAEM through a series of experiments and parameter sensitivity analyses. Concluding remarks and discussions on potential avenues for future research are presented in Section V.

\section{Related work}

Recent advances in evolutionary computation have focused on \textcolor{black}{knowledge transfer} and \textcolor{black}{online adaptation}, with ESTO and LEAs emerging as key research frontiers.

\subsection{\textcolor{black}{Evolutionary sequential transfer optimization}}

\textcolor{black}{ESTO} exploits cross-task prior knowledge to enhance the transferability of EAs in target domains, achieving notable results in hyperparameter tuning \cite{14}, neural architecture search \cite{15}, and vehicle routing \cite{9194319}. Unlike traditional trial-and-error approaches that lack prior knowledge, \textcolor{black}{ESTO} \cite{21} emphasizes systematic knowledge reuse, addressing three complementary parts: what to transfer, how to transfer, and when to transfer.

\subsubsection{What to transfer}
This involves identifying the most promising knowledge and its carriers. Solutions, as general transferable entities independent of optimizers, are gaining increasing attention. Metrics such as Hamming distance \cite{1324694}, Wasserstein distance \cite{8752421}, and ordinal correlation \cite{8401802} are used to evaluate solution similarity for selective transfer. However, current practices often underutilize prior knowledge by transferring only a limited number of solutions.

\subsubsection{How to transfer}
To enhance the quality of transferable knowledge, state-of-the-art approaches focus on learning cross-task feature mappings to predict optimal solutions for target tasks. Relevant techniques include affine transformations \cite{9295394}, linear mappings \cite{7969454}, neural networks \cite{9377001}, and so on.

\subsubsection{When to transfer}
The timing of knowledge transfer is crucial. Fixed intervals \cite{7879282} offer simplicity but lack flexibility, potentially resulting in negative transfer. In contrast, adaptive strategies \cite{8672822,CAI2021874} dynamically assess knowledge quality to determine optimal transfer times, though their effectiveness depends on the quantity and quality of available knowledge.

\subsubsection{\textcolor{black}{Theoretical analysis and benchmarks}}
\textcolor{black}{Theoretical foundations and evaluation benchmarks have been established to rigorously assess ESTO. Scott et al. \cite{12} derived the first complexity results and proved the No-Free-Lunch theorems for transfer optimization. Xue et al. \cite{xue2025theoretical} established the theoretical analysis foundation for analogy-based transfer methods. Regarding evaluation standards, Scott et al. \cite{10612149} highlighted the limitations of existing narrow benchmarks and illustrated the continuum of difficulty for knowledge reuse. To support rigorous assessment,} Xue et al. \cite{21} introduced synthetic benchmarks for evaluating ESTO methods, termed sequential transfer optimization problems (STOPs), which are characterized by fairness, reproducibility, and verifiability.

\subsection{Learnable evolutionary algorithms}

LEAs strategically integrate machine learning paradigms to learn parameter configurations within EAs to improve performance and adaptation. 
Current implementations adopt two principal learning architectures: 1) \textcolor{black}{Global methods typically employ advanced sequence modeling architectures such as recurrent neural networks (RNNs) \cite{10.1007/978-3-030-46147-8_22,NEURIPS2019_ec04e8eb} and Transformers \cite{hong2023pre,li2023b2opt,NEURIPS2024_19e9a88d,han2025enhancing} to supplant entire evolutionary frameworks.} 2) Local optimization strategies emphasize the dynamic adaptation of crucial hyperparameters \cite{shala2020learning,19,20}, such as the mutation step-size parameters of CMAES \cite{shala2020learning}. However, existing LEAs have fixed parameter configurations after training, lacking the capacity for dynamic adjustment during the evolutionary optimization process. This limits their adaptability to new scenarios.

\textcolor{black}{Meta black-box optimization (MetaBBO) has become a promising area for LEAs \cite{10993463,ma2025metabox}. MetaBBO aims to automate the design of optimizers to replace handcrafted heuristics by leveraging meta-learning techniques. For instance, reinforcement learning has been effectively utilized to automatically learn landscape features and adapt evolutionary operators \cite{10.1145/3712256.3726309}. While both MetaBBO and OKAEM share the underlying paradigm of learning from optimization progress, their core motivations diverge significantly. MetaBBO methods are typically trained across predefined task distributions to pursue zero-shot generalization on unseen tasks, without explicitly modeling the knowledge from specific source tasks. In contrast, OKAEM is specifically designed for ESTO, where the primary goal is to explicitly transfer prior knowledge across related optimization tasks to facilitate online adaptation.}

\textcolor{black}{Recently, ABOM \cite{wang2026taskfree} parameterized evolutionary operators via standard attention mechanisms. However, it is designed for MetaBBO to eliminate the reliance on predefined task distributions.ABOM adopts a task-free and single-phase mechanism for zero-shot adaptive optimization without utilizing prior knowledge. In contrast, OKAEM focuses on ESTO to simultaneously achieve knowledge transfer and online adaptation. To effectively utilize prior knowledge, OKAEM requires a more expressive architecture. Therefore, it employs multi-head attention and learnable projections within a two-phase paradigm (pre-training and adaptive optimization). While ABOM explores zero-shot optimization, OKAEM emphasizes updating parameters based on available prior knowledge.}

Despite the advancements, \textcolor{black}{ESTO} and LEAs continue to face significant challenges in effective knowledge utilization and flexible online adaptive strategies. To address these limitations, we introduce OKAEM, a neural representation-based evolutionary framework. This framework learns evolutionary behaviors from diverse prior knowledge and enables dynamic adaptation to novel scenarios, thereby achieving substantial gains in optimization efficiency.

\section{Optimization knowledge adaptation evolutionary model}

We consider applying optimization knowledge to EAs to find the optimal solution of \textcolor{black}{the target optimization task}:

\begin{equation}
    \min_{x} [f(x) \mid M, P],
\end{equation}
where \textcolor{black}{$ f\in \mathbb{R} $ }is the objective function and $ x \in \mathbb{R}^d $ is the decision variable. \textcolor{black}{
Following the existing definition of \textcolor{black}{ESTO} \cite{21}, source tasks are used to generate prior knowledge.} \textcolor{black}{$ M = \{(P_k^{(t)}, F_k^{(t)}) \mid P_k^{(t)} \in \mathbb{R}^{N \times d}, F_k^{(t)} \in \mathbb{R}^{N}, k=1,\ldots,K, t=1,\ldots,T\} $} \textcolor{black}{represents a series of prior knowledge generated by various EAs across $K$ source tasks.}
$ P_k^{(t)} $ and $ F_k^{(t)} $ denote the population and fitness data of the \textit{k}-th source task at generation $ t $. Each individual in the population represents a candidate solution for the optimization task. Fig. \ref{fig1}a visualizes the prior knowledge $ M $ accumulated by GA and CMAES on two 2D source optimization tasks: Ackley and Sphere. These visualizations intuitively reflect the evolutionary behaviors of populations on the source tasks. \textcolor{black}{$ P = \{(P^{(t)}, F^{(t)}) \mid P^{(t)} \in \mathbb{R}^{N \times d}, F^{(t)} \in \mathbb{R}^{N}, t=1,\ldots,j\} $ denotes the real-time knowledge from the 1st to the $ j $-th generation on the target task. More detailed explanations of terminology can be found in Supplementary Appendix Table VI.}

The proposed OKAEM includes two phases: 1) pre-training on $ M $ (Fig. \ref{fig1}a); adaptive optimization using $ P $ (Fig. \ref{fig1}b). The pseudocode of OKAEM is shown in Algorithm \ref{algo1}.

\begin{algorithm*}[htbp]
\caption{Optimization knowledge adaptation evolutionary model}\label{algo1}
\begin{algorithmic}[1]
\Require 
    \Statex Optimization task $\min_{x\in X} f(x), X \in \mathbb{R}^{d}$;
    \Statex Prior knowledge \textcolor{black}{$M = \{(P_k^{(t)}, F_k^{(t)}) \mid P_k^{(t)} \in \mathbb{R}^{N \times d}, F_k^{(t)} \in \mathbb{R}^{N}, k=1,\ldots,K, t=1,\ldots,T\}$ }(Optional);
    \Statex Pre-training loss $L_1$ (Optional);
    \Statex Self-tuning loss $L_2$;
    \Statex Architecture parameters: number of layers $L$, embedding dimension $d_A$, attention heads $H$, MLP hidden dimension $d_M$, dropout probability in crossover $p_C$, dropout probability in mutation $p_M$;
    \Statex Training parameters: learning rate $lr$, weight decay $wd$;
    \Statex Evolutionary parameters: population size $N$, number of iterations $T$.
\Ensure 
    \Statex Optimal individual $p^* \in \mathbb{R}^{d }$.
\State OKAEM$_W \leftarrow$ Initialization($L$, $d_A$, $H$, $d_M$, $p_C$, $p_M$); \Comment{Initialize learnable parameters}
\State $W^{(0)} \leftarrow$ AdamW($L_1$($M$), $lr$, $wd$); \Comment{Pre-train using AdamW with loss $L_1$ (Optional)}
\State $P^{(0)} \in \mathbb{R}^{N \times d} \leftarrow$ LHSampling($X$); \Comment{Initialize the population using Latin hypercube sampling}
\State \textcolor{black}{$F^{(0)} \in \mathbb{R}^{N} \leftarrow f(P^{(0)})$}; \Comment{Evaluate the fitness based on the optimization task}
\For{$t = 0$ to $T$}
    \State $\hat{P}^{(t)} \leftarrow$ $\text{OKAEM}_{W^{(t)}}(P^{(t)}, F^{(t)})$; \Comment{Reproduce using $\text{OKAEM}_{W^{(t)}}$}
    \State $\hat{F}^{(t)} \leftarrow f(\hat{P}^{(t)})$; \Comment{Evaluate the fitness based on the optimization task}
    \State $P^{*(t)}, F^{*(t)} \leftarrow$ Elitism($P^{(t)} \cup \hat{P}^{(t)}$, $F^{(t)} \cup \hat{F}^{(t)}$); \Comment{Select top $N$ individuals based on fitness}
    \State $W^{(t)} \leftarrow$ AdamW($L_2$($\hat{P}^{(t)}$, $P^{*(t)}$), $lr$, $wd$); \Comment{Self-tune using AdamW with loss $L_2$}
    \State $P^{(t+1)}, F^{(t+1)} \leftarrow P^{*(t)}, F^{*(t)}$; \Comment{Update the population and fitness}
\EndFor
\State $p^* \leftarrow$ Selection($P^{(T)}$, $F^{(T)}$); \Comment{Select the best individual from the final population}
\end{algorithmic}
\end{algorithm*}

\subsection{Architecture}
Given the current population $ P^{(t)} \in \mathbb{R}^{N \times d} $ and fitness data \textcolor{black}{$ F^{(t)} \in \mathbb{R}^{N} $}, OKAEM generates an offspring population $ \hat{P}^{(t)} \in \mathbb{R}^{N \times d} $ through selection, crossover, and mutation, denoted as $ \hat{P}^{(t)} = \text{OKAEM}_W(P^{(t)}, F^{(t)}) $ (Fig. \ref{fig1} b(i)). The selection module defines a selection matrix $ A^{(t)} \in \mathbb{R}^{N \times N} $, where $ A_{ij} $ indicates individual $ j $'s contribution to generating the $i$-th individual for crossover. Multi-head attention mechanisms \cite{NIPS2017_3f5ee243} parameterize $ A^{(t)} $ to model individual and fitness relationships. Using $ A^{(t)} $, the crossover module recombines individuals in $ P^{(t)} $ to produce an intermediate population $ P^{'(t)} $. In addition, the MLP with Dropout is employed to ensure the randomness of the crossover operator. The mutation module perturbs each $ p_i^{(t)} \in \mathbb{R}^{d} $ in $ P^{'(t)} $ to generate the corresponding offspring $ \hat{p}_i^{(t)} \in \mathbb{R}^{d} $. A mutation matrix $ M^{(t)} \in \mathbb{R}^{d \times d} $, where $ M_{jk} $ indicates the influence of the $ k $-th gene on the mutation of the $ j $-th gene, is parameterized using attention mechanisms to model gene interactions. By applying $ M^{(t)} $ to each individual $ p_i^{(t)} $, we introduce gene-level perturbations to generate each offspring $ \hat{p}_i^{(t)} $. Collectively, these offspring form the offspring population $ \hat{P}^{(t)} $. Each component are detailed as follows.

\subsubsection{Selection}

\textcolor{black}{Attention mechanisms enable adaptive modeling of interactions by computing weighted relationships between elements. In OKAEM, attention quantifies pairwise influences among individuals ($A_P^{(t)}$) or fitness scores ($A_F^{(t)}$) through learnable transformations. This allows the selection matrix to dynamically prioritize relevant individuals for crossover.}

We employ attention mechanisms to model individual and fitness relations, parameterizing the selection module. Given the current population $ P^{(t)} \in \mathbb{R}^{N \times d} $ and fitness \textcolor{black}{ $ F^{(t)} \in \mathbb{R}^{N } $}, the selection matrix $ A^{(t)} \in \mathbb{R}^{N \times N} $ is defined as follows:

\begin{equation}
    A^{(t)} = \mathrm{softmax}\left(\frac{A_P^{(t)} + A_F^{(t)}}{\sqrt{d_A}}\right). \label{eq:selection_matrix}
\end{equation}

$A_P^{(t)} = (P^{(t)} W^{QP})(P^{(t)} W^{KP})^T$ captures pairwise interactions between individuals in the population, while $A_F^{(t)} = (F^{(t)} W^{QF})(F^{(t)} W^{KF})^T$ models the interplay between their fitness scores. The matrices $ W^{QP}, W^{KP} \in \mathbb{R}^{d \times d_A} $ and $ W^{QF}, W^{KF} \in \mathbb{R}^{1 \times d_A} $ are learnable parameters that transform the original features into a space conducive to learning the selection policy. \textcolor{black}{$d_A$ represents the embedding dimension that determines the feature space size.}
The softmax function converts the relation scores into a probability distribution, scaled by $ \sqrt{d_A} $ to stabilize optimization and prevent gradient explosion. Element $ A_{ij}^{(t)} $ in the selection matrix can be used to quantify the extent to which individual $j$ influences the generation of individual $i$ by crossover.

\subsubsection{Crossover}
The crossover module begins by applying the selection matrix $ A^{(t)} $ to the current population $ P^{(t)} $ for individual-level recombination:
\begin{equation}\label{eq:OC}
    O_C^{(t)} = A^{(t)} P^{(t)} W^V,
\end{equation}
where the transformation $ W^V \in \mathbb{R}^{d \times d_A}$ reduces the dimensionality of the search space.  This dimensionality reduction is a well-established technique in EAs to enhance optimization efficiency and reduce computational complexity \cite{10.5555/3060832.3060893,10.1145/3470971}.

To enhance the expressiveness of the crossover, we employ multi-head attention mechanisms that focus on the features of the population and fitness across different subspaces. Given $ H $ heads, Eq.~\eqref{eq:OC} becomes:
\begin{equation}\label{OCH}
    O_C^{(t)} = \|_{h=1}^H A_h^{(t)} P^{(t)} W_h^V,
\end{equation}
where
\begin{equation}
    \begin{aligned}
    A_h^{(t)} = \text{Softmax} & \left(
    \frac{(P^{(t)} W_h^{QP})(P^{(t)} W_h^{KP})^T}{\sqrt{d_A / H}} \right. \\
    & \left. + \frac{(F^{(t)} W_h^{QF})(F^{(t)} W_h^{KF})^T}{\sqrt{d_A / H}} \right),
    \end{aligned}
\end{equation}

and $ W_h^V, W_h^{QP}, W_h^{KP} \in \mathbb{R}^{d \times (d_A / H)} $, $ W_h^{QF}, W_h^{KF} \in \mathbb{R}^{1 \times (d_A / H)} $. \textcolor{black}{The symbol $\|_{h=1}^H$ denotes concatenation across attention heads.}

Next, we introduce nonlinearity into $ O_C^{(t)} $ using a MLP with Dropout:
\begin{equation}\label{mlpc}
    MLP(O_C^{(t)}) = \text{Dropout}_{p_C}(\text{Tanh}(O_C^{(t)} W_1 + b_1)) W_2 + b_2,
\end{equation}
where $ \text{Tanh}() $ is the activation function, and $ W_1 \in \mathbb{R}^{d_A \times d_M} $, $ b_1 \in \mathbb{R}^{d_M} $, $ W_2 \in \mathbb{R}^{d_M \times d} $, $ b_2 \in \mathbb{R}^{d} $ are parameters. \textcolor{black}{$d_M$ denotes the hidden layer dimension of MLPs.} The Dropout introduces stochasticity, with probability $ p_C $ controlling neuron deactivation, thereby modulating the randomness of the crossover process. 
\textcolor{black}{To ensure stochastic behavior while maintaining learnability, we apply Dropout during both training and inference. During reproduction, random neuron subsets in the crossover and mutation modules are masked, creating dynamic pathways for diverse offspring generation. This contrasts with fixed-parameter LEAs that lose randomness post-training.}
Finally, residual connections are used to generate the post-crossover intermediate population:
\begin{equation}
    P^{'(t)} = P^{(t)} + MLP(O_C^{(t)}),
\end{equation}
which helps mitigate gradient vanishing during training, ensuring stable convergence.

In summary, the learnable parameters $ W_{SC} $ for the selection and crossover modules are:
\begin{equation}
\begin{aligned}
        W_{SC} = \{&W_h^{QP}, W_h^{KP}, W_h^{QF}, W_h^{KF}, W_h^V, \\
        & W_1, b_1, W_2, b_2, h = 1, ..., H\}.
\end{aligned}
\end{equation}

\subsubsection{Mutation}
The mutation module individually perturbs each individual $ p_i^{'(t)} \in \mathbb{R}^{d} $ in the intermediate population $ P'^{(t)} $ to generate the corresponding offspring $ \hat{p}_i^{(t)} \in \mathbb{R}^{d} $. Employing attention mechanisms, we model gene relations to guide the mutation process. The mutation matrix $ M^{(t)} \in \mathbb{R}^{d \times d} $ is defined as:

\begin{equation}\label{eq:M}
    M^{(t)} = \mathrm{Softmax}\left(\frac{(p_i^{'(t)} W^{QM}) (p_i^{'(t)} W^{KM})^T}{\sqrt{d_A}}\right),
\end{equation}
where $ W^{QM}, W^{KM} \in \mathbb{R}^{1 \times d_A} $. Similar to the crossover module, we use residual connections and a MLP to generate offspring:
\begin{equation}\label{eq:offspring}
    \hat{p}_i^{(t)} = p_i^{'(t)} + MLP(o_{M,i}^{(t)}),
\end{equation}
with
\begin{equation}\label{eq:OMT}
    o_{M,i}^{(t)} = M^{(t)} p_i^{'(t)} W^{VM},
\end{equation}
and
\begin{equation}\label{eq:MLPM}
    MLP(o_{M,i}^{(t)}) = \mathrm{Dropout}_{p_M}(\mathrm{Tanh}(o_{M,i}^{(t)} W_3 + b_3)) W_4 + b_4,
\end{equation}
where $ W^{VM} \in \mathbb{R}^{1 \times d_A} $, $ W_3 \in \mathbb{R}^{d_A \times d_M} $, $ b_3 \in \mathbb{R}^{d_M} $, $ W_4 \in \mathbb{R}^{d_M} $, and $ b_4 \in \mathbb{R} $. The Dropout layer with probability $ p_M $ introduces stochasticity into the mutation process.

All generated offspring are then combined into the offspring population $ \hat{P}(t) \in \mathbb{R}^{N \times d} $:
\begin{equation}\label{eq:offspring_population}
    \hat{P}(t) = \|_{i=1}^N (\hat{p}_i^{(t)})^T,
\end{equation}
\textcolor{black}{where the symbol $\|_{i=1}^N$ denotes concatenation of the $N$ offspring individuals.}

The parameters of the mutation module are summarized as:
\begin{equation}\label{eq:WM}
    W_M = \{W^{QM}, W^{KM}, W^{VM}, W_3, b_3, W_4, b_4\}.
\end{equation}

In summary, the learnable parameters $ W $ of a single layer of OKAEM include all weight matrices and bias vectors from the selection, crossover, and mutation:
\begin{equation}\label{eq:all_params}
\begin{aligned}
    W = \{&W_{SC}, W_M\} \\
    = \{&W_h^{QP}, W_h^{KP}, W_h^{QF}, W_h^{KF}, W_h^V, W_1, b_1, W_2, b_2,\\
    &W^{QM}, W^{KM}, W^{VM}, W_3, b_3, W_4, b_4, h = 1, ..., H\}.
\end{aligned}
\end{equation}

\subsubsection{Discussion}

Compared to traditional EAs, OKAEM offers three key advantages: learnability, parallelism, and interpretability.

\begin{itemize}
    \item \textbf{Learnability}: The parameterized selection, crossover, and mutation modules can be adaptively updated based on optimization knowledge, rather than relying on heuristic rules. This adaptive learning capability enhances the generalization performance.
    
    \item \textbf{Parallelism}: By adhering to the fundamental design principles of neural networks, OKAEM enables the direct application of existing GPU-based parallel computing strategies. This significantly reduces computational costs and accelerates the optimization process.
    
    \item \textbf{Interpretability}: OKAEM allows for the visualization of the selection and mutation matrices, providing valuable insights into individual relationships, fitness dynamics, and gene interactions during the evolutionary process. Specifically, observations from these visualizations reveal distinct statistical patterns: individuals with higher fitness are more likely to be selected for crossover, and gene mutations exhibit consistent patterns. For a detailed analysis of these advantages, see Fig \ref{fig3}.
\end{itemize}

\subsection{Pre-training and adaptive optimization}
The pre-training aims to uncover the patterns underlying prior knowledge to enhance the performance of OKAEM. As shown in Fig.~\ref{fig1}a, given prior knowledge $M$, the optimization objective of pre-training is to minimize the Euclidean distance between the predicted offspring population $ \hat{P}_k^{(t)} $ and the actual next-generation population $ P_k^{(t+1)} $ in the prior knowledge:
\begin{equation}
    \begin{aligned}
        \min_W \quad & L_1 = \sum_{k=1}^K \sum_{t=1}^{T-1} \| P_k^{(t+1)} - \hat{P}_k^{(t)} \|^2, \\ &\hat{P}_k^{(t)} = \text{OKAEM}_W(P_k^{(t)}, F_k^{(t)}).
    \label{eq:pretrain_objective}
    \end{aligned}
\end{equation}

This allows OKAEM to explicitly learn population evolution behavior by predicting the next generation. Traditional \textcolor{black}{ESTO} methods rely on heuristic rules to determine what, when, and how to transfer knowledge, focusing only on a subset of promising prior knowledge and heavily depending on expert design \cite{11,21}. In contrast, OKAEM utilizes all prior knowledge for pre-training, avoiding knowledge waste. Moreover, prior knowledge generated by different EAs on a set of source optimization tasks can be used as training data, enabling OKAEM to learn diverse types of evolution behaviors (see detailed analysis in Fig.~\ref{fig2}b).

As shown in Fig.~\ref{fig1}b, adaptive optimization comprises initialization, reproduction, evaluation, elitism, and self-tuning. Over $ T $ generations, the best individual $ p^* \in \mathbb{R}^{d} $ is obtained. Initially, Latin hypercube sampling is used to generate a random initial population from the search space. By leveraging the pre-trained OKAEM, we proceed with population reproduction. Subsequently, the offspring population undergoes fitness evaluation. Furthermore, an elitism strategy ensures that the top $ N $ individuals with the highest fitness are carried over to the next generation. Finally, self-tuning updates OKAEM's parameters by minimizing the distance between the generated offspring population $ \hat{P}^{(t)} $ and the elite individuals $ P^{*(t)} $:
\begin{equation}
    \min_W L_2 = \| P^{*(t)} - \hat{P}^{(t)} \|^2.
    \label{eq:fine_tune_objective}
\end{equation}

During the self-tuning phase, OKAEM iteratively updates its parameters to adapt to the current target task. From a learning perspective, pre-training and self-tuning correspond to unsupervised and supervised learning paradigms, respectively. We can use classical gradient optimizers such as AdamW \cite{loshchilov2018decoupled} to update the model parameters $ W $. Pre-training aims to more comprehensively utilize prior knowledge (transferability), while self-tuning aims to learn new knowledge generated by itself (adaptability). These processes leverage population dynamics and fitness evaluations, independent of explicit gradient information from the fitness function, providing a robust training strategy for EAs. \textcolor{black}{Furthermore, the modular design of pre-training and self-tuning ensures effectiveness even without prior knowledge, enabling direct adaptive optimization on the target task. This dual-phase approach enhances OKAEM's ability to generalize and adapt, making it suitable for a wide range of optimization challenges.}

\subsection{Computational complexity analysis}

The selection matrix and the MLP primarily determine the computational complexity of the selection and crossover. According to Eq. (\ref{eq:selection_matrix}), the complexity of computing the selection matrix is $ O(N \cdot d \cdot d_A + N^2 \cdot d_A) $, where $ N $ is the population size, $ d $ is the dimensionality of the search space, and $ d_A $ is the embedding dimension. From Eq. (\ref{mlpc}), the complexity of the MLP involved in the crossover is $ O(N \cdot d_A \cdot d_M + N \cdot d_M \cdot d) $, with $ d_M $ representing the hidden layer dimension. Therefore, the total complexity for selection and crossover is $
O(N \cdot d \cdot d_A + N^2 \cdot d_A + N \cdot d_A \cdot d_M + N \cdot d_M \cdot d)$.

According to Eq. (\ref{eq:M}), the complexity of computing the mutation matrix is $ O(d \cdot d_A + d^2 \cdot d_A) $. Eq. (\ref{eq:MLPM}) indicates that the MLP involved in mutation has a complexity of $ O(d \cdot d_A \cdot d_M) $. Thus, the overall complexity for mutation is: $ O(d^2 \cdot d_A + d \cdot d_A \cdot d_M) $.

Consolidating these results, the total computational complexity for an $ L $-layer architecture is $ O(L \cdot N \cdot d \cdot d_A + L \cdot N^2 \cdot d_A + L \cdot N \cdot d_A \cdot d_M + L \cdot N \cdot d_M \cdot d + L \cdot d^2 \cdot d_A + L \cdot d \cdot d_A \cdot d_M) $. Assuming $ d_A = d_M = d $, this simplifies to $ O(L \cdot N \cdot d^2 + L \cdot N^2 \cdot d + L \cdot d^3) $. \textcolor{black}{In complex optimization scenarios, where the population size $ N $ is generally smaller than the problem dimension $ d $, the leading term of complexity is $ O(L \cdot d^3) $. This highlights the significant impact of the problem dimension $ d $ on computational requirements in high-dimensional search spaces. The assumption $ d_A = d_M = d $ simplifies the analysis but may not hold in practice. Careful consideration should be given to the specific values of $ d_A $ and $ d_M $ based on the application context.}

\subsection{\textcolor{black}{Convergence analysis of adaptive optimization in OKAEM}}\label{appendix:convergence}
\textcolor{black}{This section establishes the theoretical foundation for the global convergence of OKAEM's adaptive optimization phase. By modeling the population dynamics as a stochastic process, we prove that the neural parameterized evolutionary operators, combined with the elitism strategy, ensure almost sure convergence to the global optimum.}

\textcolor{black}{Let $\mathcal{X} \subset \mathbb{R}^d$ be a compact search space. The objective function $f: \mathcal{X} \to \mathbb{R}$ is assumed to be continuous, with a global minimum $x^* \in \mathcal{X}$. To facilitate the theoretical analysis, we postulate the following standard assumptions regarding the optimization landscape and the neural parameterized evolutionary operators:}

\begin{assumption}\label{as:app}
\textcolor{black}{It is assumed that:}
\begin{enumerate}
    \item[(1)] \textcolor{black}{The global optimum lies in the interior of the search space, i.e., $x^* \in \mathrm{int}(\mathcal{X})$;}
    \item[(2)] \textcolor{black}{MLP with non-linear activation functions (e.g., $\mathrm{Tanh}$) has a hidden dimension $d_M \geq 1$;}
    \item[(3)] \textcolor{black}{The dropout rate satisfies $0 < p_C, p_M < 1$.}
\end{enumerate}
\end{assumption}

\textcolor{black}{We first examine the stability of the optimization process. Let $f^*_t = \min_{p \in P^{*(t)}} f(p)$ be the best objective value in the elite archive $P^{*(t)}$ at generation $t$. The following lemma guarantees that the quality of the best-found solution does not degrade over time. }

\begin{lemma}[Monotonicity]\label{lem:monotonicity}
\textcolor{black}{The sequence of the best objective values $\{f_t^*\}_{t=0}^{\infty}$ is monotonically non-increasing and converges almost surely to a random variable $f_\infty^*$.}
\end{lemma}

\begin{proof}
\textcolor{black}{According to the elitism strategy in Algorithm \ref{algo1}, the elite archive $P^{*(t+1)}$ retains the best individuals from the union of the current population and the offspring. Consequently, $f_{t+1}^* \le f_t^*$ holds for all $t$. Since $\mathcal{X}$ is compact and $f$ is continuous, the sequence $\{f_t^*\}$ is lower-bounded by $f^*$. By the monotone convergence theorem \cite{hall2014martingale}, the sequence converges almost surely to a limit $f_\infty^* \ge f^*$.}
\end{proof}

\textcolor{black}{While monotonicity ensures stability, global convergence requires the capability to explore the entire search space. \textcolor{black}{Let $\mathcal{F}_t$ be the filtration generated by the sequence of populations $P$ and parameters $W$ up to generation $t$, formally defined as the $\sigma$-algebra $\mathcal{F}_t = \sigma(P^{(0)}, \dots, P^{(t)}, W^{(0)}, \dots, W^{(t)})$. This captures the complete history of the algorithm's state.} The next lemma quantifies the exploration capability of the proposed OKAEM.}

\begin{lemma}[Exploration Capability]\label{lem:exploration}
\textcolor{black}{Under Assumption \ref{as:app}, for any $\delta > 0$, there exists $\gamma > 0$ such that the probability of generating an offspring in the $\delta$-neighborhood of $x^*$ is strictly positive:}
\begin{equation}\label{eq:exploration}
\textcolor{black}{\mathbb{P}(\exists i: \|\hat{p}_i^{(t)} - x^*\| < \delta \mid \mathcal{F}_t) \geq 1 - (1-\gamma)^N > 0.}
\end{equation}
\end{lemma}

\begin{proof}
\textcolor{black}{We analyze evolutionary operators of OKAEM sequentially. Let $W_{SC}$ and $W_M$ denote the parameters for the crossover (with selection) and mutation, respectively, as defined in Eq. (\ref{eq:all_params}).}

\textcolor{black}{1) Crossover with selection: Consider any parent individual $p_i^{(t)}$ (the $i$-th row of $P^{(t)}$). The crossover with selection is:}
\begin{equation}
\textcolor{black}{p_i^{'(t)} = p_i^{(t)} + \mathrm{MLP}\left(\left[O_C^{(t)}\right]_i\right),}
\end{equation}
\textcolor{black}{where $\left[O_C^{(t)}\right]_i=\left[\ \|_{h=1}^H A_h^{(t)} P^{(t)} W_h^V \right]_i$ represents the $i$-th row of the concatenated multi-head attention output, aggregating global population information.}

\textcolor{black}{Let $v = x^* - p_i^{(t)}$ be the target displacement. We construct a feasible reference configuration $W_{SC}^*$ by setting the MLP parameters $W_1 = W_2 = b_1 = 0$, and the output bias $b_2 = v$:}
\begin{equation}
\begin{split}
\textcolor{black}{\mathrm{MLP}(\left[\dots\right]_i)} & \textcolor{black}{= \mathrm{Dropout}_{p_C}\left(\mathrm{Tanh}\left((\left[\dots\right]_i  W_1 + b_1\right)\right) W_2 + b_2 } \\
& \textcolor{black}{= \mathrm{Dropout}_{p_C}(\mathrm{Tanh}(0)) \cdot 0 + v = v.}
\end{split}
\end{equation}

\textcolor{black}{Thus, $p_i^{'(t)} = p_i^{(t)} + v = x^*$.}
\textcolor{black}{By the continuity of the MLP with respect to its parameters, there exists $\epsilon > 0$ defining an open ball $\mathcal{B}_\epsilon(W_{SC}^*) = \{W : \|W - W_{SC}^*\| < \epsilon\}$ such that for all $W_{SC} \in \mathcal{B}_\epsilon(W_{SC}^*)$, the condition $\|p_i^{'(t)} - x^*\| < \delta/2$ holds.}

\textcolor{black}{The parameters update via $W_{SC}^{(t+1)} = W_{SC}^{(t)} - \eta \nabla L_2 + \xi^{(t)}$, where the stochastic perturbation $\xi^{(t)}$ depends on the dropout mask $D^{(t)}$. The minimum probability mass is:} 
\begin{equation}
\textcolor{black}{\min_{D} \mathbb{P}(D^{(t)} = D) = (\min\{p_C, 1-p_C\})^{d_M} > 0. }
\end{equation}

\textcolor{black}{Therefore, there exists $c_t > 0$ such that:}
\begin{equation}
\begin{split}
\textcolor{black}{\mu_c} & \textcolor{black}{= \mathbb{P}(W_{SC}^{(t)} \in \mathcal{B}_\epsilon(W_{SC}^*) \mid \mathcal{F}_t)} \\
& \textcolor{black}{\geq (\min\{p_C, 1-p_C\})^{d_M} \cdot c_t > 0.}
\end{split}
\end{equation}

\textcolor{black}{2) Mutation: Similarly, consider the mutation update for the intermediate individual $p_i^{'(t)}$. The update rule is:}
\begin{equation}
\textcolor{black}{\hat{p}_i^{(t)} = p_i^{'(t)} + \mathrm{MLP}\left(o_{M,i}^{(t)}\right),}
\end{equation}
\textcolor{black}{where $o_{M,i}^{(t)} = M^{(t)} p_i^{'(t)} W^{VM}$ denotes the mutation feature vector for the $i$-th individual.}

\textcolor{black}{Let $w = x^* - p_i^{'(t)}$. We identify a candidate configuration $W_M^*$ by setting $W_3 = W_4 = b_3 = 0$ and $b_4 = w$. This yields:}
\begin{equation}
\textcolor{black}{\mathrm{MLP}\left(o_{M,i}^{(t)}\right) = \mathrm{Dropout}_{p_M}(\mathrm{Tanh}(0)) \cdot 0 + w = w.}
\end{equation}

\textcolor{black}{Consequently, $\hat{p}_i^{(t)} = p_i^{'(t)} + w = x^*$. Analogous to the crossover, by continuity, there exists an open ball $\mathcal{B}_{\epsilon'}(W_M^*) = \{W : \|W - W_M^*\| < \epsilon'\}$ such that for all $W_M \in \mathcal{B}_{\epsilon'}(W_M^*)$, the offspring satisfies $\|\hat{p}_i^{(t)} - x^*\| < \delta/2$.}

\textcolor{black}{The parameters are updated via $W_M^{(t+1)} = W_M^{(t)} - \eta \nabla L_2 + \xi^{(t)}$, where the perturbation $\xi^{(t)}$ is derived from the dropout mask. The minimum probability mass is $(\min\{p_M, 1-p_M\})^{d_M} > 0$. Therefore, there exists $c'_t > 0$ such that:}
\begin{equation}
\begin{split}
\textcolor{black}{\mu_m} & \textcolor{black}{= \mathbb{P}(W_M^{(t)} \in \mathcal{B}_{\epsilon'}(W_M^*) \mid \mathcal{F}_t)} \\
& \textcolor{black}{\geq (\min\{p_M, 1-p_M\})^{d_M} \cdot c'_t > 0.}
\end{split}
\end{equation}

\textcolor{black}{3) Overall Success Probability: The probability that a single offspring $\hat{p}_i^{(t)}$ falls within the $\delta$-neighborhood of $x^*$ is lower-bounded by the joint success of the crossover and mutation:}
\begin{equation}
\textcolor{black}{\mathbb{P}(\|\hat{p}_i^{(t)} - x^*\| < \delta) \geq \mu_c \mu_m.}
\end{equation}

\textcolor{black}{Let $\gamma = \mu_c \mu_m > 0$ denote the success probability for a single individual. Considering $N$ independent offspring generated in parallel, the probability that at least one offspring reaches the target region is:}
\begin{equation}
\begin{split}
\textcolor{black}{\mathbb{P}(\exists i: \|\hat{p}_i^{(t)} - x^*\| < \delta)} & \textcolor{black}{= 1 - \prod_{i=1}^N \left(1 - \mathbb{P}(\|\hat{p}_i^{(t)} - x^*\| < \delta)\right)} \\
& \textcolor{black}{\geq 1 - (1 - \gamma)^N > 0.}
\end{split}
\end{equation}

\end{proof}

\begin{remark}
\textcolor{black}{Lemma \ref{lem:exploration} relies on constructing feasible reference configurations $W_{SC}^*$ and $W_M^*$ to demonstrate the theoretical reachability of the optimal region. It is worth noting that in practice, the exploration is driven by the interplay between the self-tuning loss $L_2$ and the dropout mechanism. The structural stochasticity induced by dropout ensures that the evolutionary operators maintain a non-zero probability support over the parameter space. Crucially, even if the gradient guidance from $L_2$ becomes uninformative, this intrinsic stochasticity guarantees that the operators retain the capacity to probe the neighborhood of $x^*$, securing the exploration condition required for global convergence.}
\end{remark}

\textcolor{black}{Define the optimality gap $V_t = f^*_t - f^* \geq 0$. Using the exploration capability from Lemma \ref{lem:exploration}, we establish the following drift condition \cite{he2001drift,zhou2019evolutionary}:}

\begin{lemma}[Drift Condition]\label{lem:drift_app}
\textcolor{black}{Under Assumption \ref{as:app}, for any $\epsilon > 0$, there exists $\eta(\epsilon) > 0$ such that:}
\begin{equation}\label{eq:drift_condition}
\textcolor{black}{\mathbb{E}[V_t - V_{t+1} \mid \mathcal{F}_t, V_t > \epsilon] \geq \eta(\epsilon) > 0.}
\end{equation}
\end{lemma}

\begin{proof}
\textcolor{black}{By continuity and Assumption \ref{as:app} (1), there exists $\delta > 0$ such that $\|x - x^*\| < \delta$ implies $f(x) < f^* + \epsilon/2$. Define the success event $A_t = \{\exists i: \|\hat{p}_i^{(t)} - x^*\| < \delta\}$. From Lemma \ref{lem:exploration}, its probability satisfies:}
\begin{equation}\label{eq:prob_A}
\textcolor{black}{\mathbb{P}(A_t \mid \mathcal{F}_t, V_t > \epsilon) \geq 1 - (1-\gamma)^N := \nu > 0.}
\end{equation}

\textcolor{black}{The expected drift decomposes as:}
\begin{equation}
\begin{split}
\textcolor{black}{\mathbb{E}[V_t - V_{t+1} \mid \mathcal{F}_t]} & \textcolor{black}{= \mathbb{E}[V_t - V_{t+1} \mid \mathcal{F}_t, A_t] \mathbb{P}(A_t)} \\
& \quad \textcolor{black}{+ \mathbb{E}[V_t - V_{t+1} \mid \mathcal{F}_t, A_t^c] \mathbb{P}(A_t^c).}
\end{split}
\end{equation}

\textcolor{black}{For the first term, the occurrence of $A_t$ implies that the updated archive contains a solution with fitness better than $f^* + \epsilon/2$, i.e., $V_{t+1} < \epsilon/2$. Combined with the condition $V_t > \epsilon$, we have:}
\begin{equation}
\textcolor{black}{\mathbb{E}[V_t - V_{t+1} \mid \mathcal{F}_t, A_t] > \epsilon - \epsilon/2 = \epsilon/2.}
\end{equation}

\textcolor{black}{For the second term, when $A_t$ does not occur, the monotonicity property (Lemma \ref{lem:monotonicity}) guarantees $V_{t+1} \leq V_t$. Consequently, the drift is non-negative:}
\begin{equation}
\textcolor{black}{\mathbb{E}[V_t - V_{t+1} \mid \mathcal{F}_t, A_t^c] \geq 0.}
\end{equation}

\textcolor{black}{Substituting these bounds back into the decomposition equation yields:}
\begin{equation}
\begin{split}
\textcolor{black}{\mathbb{E}[V_t - V_{t+1} \mid \mathcal{F}_t, V_t > \epsilon]} & \textcolor{black}{\geq (\epsilon/2) \cdot \nu + 0 \cdot (1-\nu)} \\
& \textcolor{black}{= (\epsilon/2)(1 - (1-\gamma)^N).}
\end{split}
\end{equation}
\textcolor{black}{The proof is completed by setting $\eta(\epsilon) = (\epsilon/2)(1 - (1-\gamma)^N)$.}
\end{proof}

\textcolor{black}{Combining the exploration capability from Lemma \ref{lem:exploration} and the drift condition provided by Lemma \ref{lem:drift_app}, we now prove the global convergence.}

\begin{theorem}[Global Convergence]\label{thm:convergence}
\textcolor{black}{Under Assumption \ref{as:app}, OKAEM's adaptive optimization phase converges to the global optimum almost surely: $f^*_t \xrightarrow{\text{a.s.}} f^*$.}
\end{theorem}

\begin{proof}
\textcolor{black}{By Lemma \ref{lem:monotonicity} and the martingale convergence theorem \cite{hall2014martingale}, $\{f^*_t\}$ converges almost surely to a random variable $f^*_\infty \geq f^*$. Assume for contradiction that $\mathbb{P}(f^*_\infty > f^*) > 0$. Then there exists $\epsilon > 0$ such that $V_t > \epsilon$ for sufficiently large $t$. Define the first hitting time $\tau_k = \inf\{t \geq k: V_t \leq \epsilon\}$.} \textcolor{black}{Using Lemma \ref{lem:drift_app}, the expected reduction in $V_t$ from time $k$ to $\tau_k$ satisfies:}
\begin{equation}
\textcolor{black}{\mathbb{E}[V_k - V_{\tau_k}] = \mathbb{E}\left[\sum_{t=k}^{\tau_k-1} \mathbb{E}[V_t - V_{t+1} \mid \mathcal{F}_t]\right] \geq \eta(\epsilon) \mathbb{E}[\tau_k - k].}
\end{equation}

\textcolor{black}{Thus:}
\begin{equation}
\textcolor{black}{\mathbb{E}[\tau_k - k] \leq \frac{\mathbb{E}[V_k] - \mathbb{E}[V_{\tau_k}]}{\eta(\epsilon)} \leq \frac{\mathbb{E}[V_k]}{\eta(\epsilon)} < \infty.}
\end{equation}

\textcolor{black}{The finiteness of the expected first hitting time implies $\mathbb{P}(\tau_k < \infty) = 1$, contradicting the hypothesis that $V_t > \epsilon$ persists indefinitely. Thus, $f^*_\infty = f^*$ almost surely.}
\end{proof}

\begin{remark}
\textcolor{black}{Based on the convergence analysis, our theoretical contributions and limitations are summarized as follows:}
\begin{itemize}
    \item \textcolor{black}{\textbf{Exploration Guarantee (Lemma \ref{lem:exploration}):} We prove that the proposed neural operators possess persistent exploration capabilities, constituting the first theoretical guarantee for LEAs. However, the current analysis assumes interior optima, excluding boundary cases or discontinuous feasible regions.}

    \item \textcolor{black}{\textbf{Global Convergence (Theorem \ref{thm:convergence}):} The global convergence of the adaptive optimization phase in OKAEM is established via drift analysis, which stands in contrast to existing ESTO and LEA methods that lack stability proofs. Note that while asymptotic convergence is ensured, deriving the convergence rate for specific problems (e.g., quadratic functions) remains an open challenge.}
\end{itemize}
\end{remark}

\section{Experiments}\label{secres}

\textcolor{black}{Section IV systematically validates OKAEM through four complementary aspects:}  

\begin{itemize}
    \item \textcolor{black}{STOP suite: Verifies the capability of using prior knowledge against established \textcolor{black}{ESTO} baselines and validates the effectiveness of pre-training and self-tuning mechanisms.}
    \item \textcolor{black}{BBOB suite: Evaluates self-tuning performance without pre-training by comparing against state-of-the-art LEAs.}
    \item \textcolor{black}{Black-box prompt tuning for vision-language models: Demonstrates real-world applicability through competitive accuracy and reduced computational cost.}
    \item \textcolor{black}{Parameter sensitivity analysis: Conducts ablation studies on critical parameters to assess robustness.}
\end{itemize}

\subsection{Sequential transfer optimization problem}

\subsubsection{Problem configuration}
We evaluate OKAEM's performance on the STOP suite \cite{21}, a comprehensive benchmark comprising 12 problems that simulate knowledge transfer scenarios in EAs \footnote{The implementations of STOPs can be accessed at \url{https://github.com/XmingHsueh/STOP-G}.}. Each STOP task encompasses a target task and prior knowledge derived from a set of source optimization tasks, effectively representing the spectrum of similarity relationships between optimal solutions encountered in real-world applications. The problems are categorized into three groups based on their degree of similarity: high (STOP1-4), mixed (STOP5-8), and low (STOP9-12). For detailed problem configurations, see Supplementary Appendix A. 

\subsubsection{Baseline}

The baselines include both classical and advanced \textcolor{black}{ESTO} methods \cite{21}, encompassing various dimensions of knowledge transfer: non-transfer strategies, what to transfer, how to transfer, and when to transfer. Detailed implementations of these methods are provided in \cite{21}.

\begin{itemize}
    \item \textbf{No knowledge transfer (N)} involves using evolutionary optimizers without incorporating any prior knowledge.
    
    \item \textbf{What to transfer} focuses on identifying solutions with the highest estimated transferability for target tasks. Evaluation metrics include:
    \begin{itemize}
        \item Hamming distance (H)
        \item Euclidean distance (M1)
        \item Wasserstein distance (WD)
        \item Ordinal correlation (OC)
        \item Relaxed ordinal correlation (ROC)
        \item Kullback-Leibler divergence (KLD)
    \end{itemize}
    
    \item \textbf{How to transfer} aims at enhancing the quality of transferred solutions. These approaches typically involve adjusting solutions using the learned mapping between the source and target tasks. Techniques include:

    \begin{itemize}
        \item Elite-based translation transformation (M1-Te)
        \item Random individual-based translation transformation (M1-Tr)
        \item Population mean-based translation transformation (M1-Tm)
        \item Multiplication transformation using estimated means (M1-M)
        \item Affine transformation (M2-A)
        \item Linear transformation with ordinal correlation (OC-L)
        \item Affine transformation with ordinal correlation (OC-A)
        \item Kernel mapping (OC-K)
        \item Neural network models (OC-N)
        \item Latent-space connected linear transformations (ROC-L)
    \end{itemize}
    
    \item \textbf{When to transfer} addresses timing decisions for knowledge transfer throughout the evolutionary process. This category includes fixed-interval methods, denoted as F1, F5, and F10, and adaptive methods such as:
    \begin{itemize}
        \item Mixture model-based estimation (D-M)
        \item Representation model-based estimation (D-G)
        \item Population distribution-based estimation (D-P)
    \end{itemize}
\end{itemize}

Additionally, two variants of OKAEM serve as baselines for comparison: OKAEM-PT, which relies solely on pre-training, and OKAEM-ST, which focuses exclusively on self-tuning. This setup allows us to comprehensively assess the effectiveness of OKAEM's dual-phase approach in leveraging prior knowledge and adapting to new tasks. 

\subsubsection{Parameter setting}

All experiments are conducted on a Linux platform with a GPU 2080Ti (Memory: 12 GB, CUDA Version: 11.3) \footnote{Matlab codes for \textcolor{black}{ESTO} methods can be found at \url{https://github.com/XmingHsueh/STO-EC}.}.\textcolor{black}{The parameter settings for all methods can be found in Supplementary Appendices A and D.}

\subsubsection{Result} 

\begin{figure*}[htbp]
\centering
\includegraphics[width=0.83\textwidth]{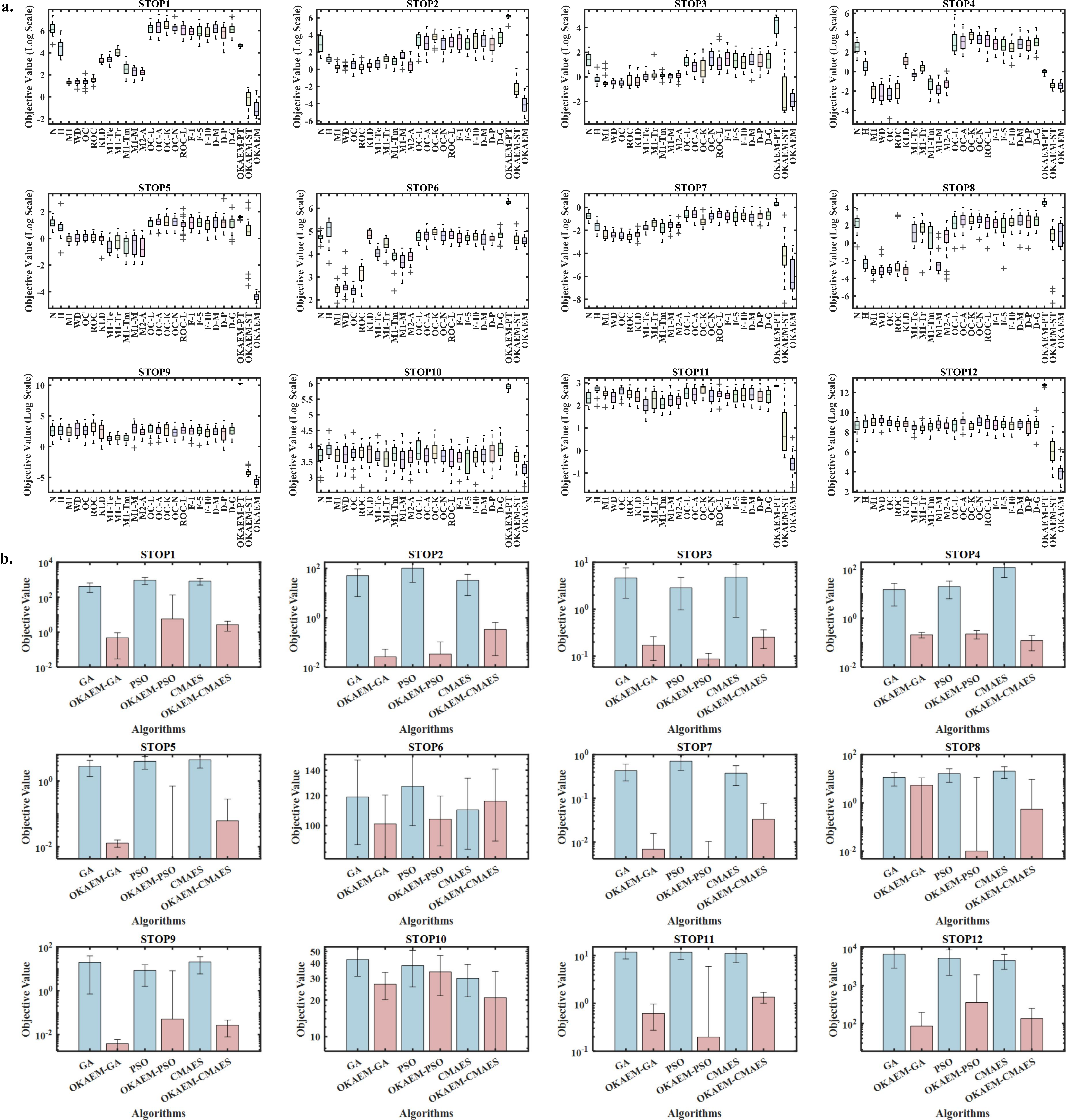}
\caption{Experimental results on the STOP suite over 20 independent runs, each with a maximum of 5000 evaluations. a. Logarithmic plot of objective values (lower is better).
b. Performance comparison between OKAEM, which learns different types of prior knowledge, and the corresponding source algorithms on target tasks. Data are presented as the mean and standard deviation of the objective values.}\label{fig2}
\end{figure*}

As shown in Fig.~\ref{fig2}a, the experimental results provide several key insights:

\begin{itemize}
    \item \textbf{Superior performance}: OKAEM demonstrates significantly better performance compared to baselines in complex knowledge transfer scenarios. \textcolor{black}{Unlike existing \textcolor{black}{ESTO} methods that focus primarily on promising individuals or their variants, OKAEM leverages comprehensive prior knowledge, including all individuals and their fitness, thereby preventing knowledge wastage.}
    \item \textbf{Effectiveness in similar tasks}:  In high or mixed similarity scenarios (STOP1-STOP8), many \textcolor{black}{ESTO} methods outperform non-transfer approaches (N), underscoring the effectiveness and necessity of knowledge transfer. Prior knowledge from source tasks markedly enhances the performance of EAs.
    \item \textbf{Resilience in low similarity}: Most \textcolor{black}{ESTO} methods perform worse than non-transfer approaches in low-similarity scenarios (STOP9-STOP12) due to limited shared knowledge and negative transfer \cite{21}. Notably, OKAEM excels even under these challenging conditions. Pre-training on low-similarity source tasks improves model generalizability, leading to superior target task performance.
    \item \textbf{Importance of pre-training and self-tuning}: \textcolor{black}{In all transfer scenarios, OKAEM outperforms both OKAEM-ST and OKAEM-PT, highlighting the critical role of integrating both pre-training and self-tuning phases in enhancing overall performance. In addition, OKAEM consistently outperforms OKAEM-PT (pre-training only, no self-tuning) across all similarity levels. In low-similarity scenarios (STOP9-STOP12), most of the source tasks exhibit negative transfer \cite{21}. OKAEM's self-tuning mechanism suppresses harmful knowledge through real-time adaptation. This is achieved by minimizing $ L_2 = \| P^{*(t)} - \hat{P}^{(t)} \|^2 $, which dynamically recalibrates model parameters based on current population-fitness interactions. Harmful knowledge may cause population drift from the target task's optimal solution, but self-tuning inhibits this by aligning offspring generation with elite individuals' traits. This autonomous adaptation mechanism enables OKAEM to maintain performance stability even when prior knowledge contains negative transfer patterns.} 
    \item \textbf{Enhanced optimization by self-tuning}: 
    \textcolor{black}{OKAEM significantly surpasses OKAEM-PT, with the key difference being the execution of self-tuning during the optimization phase. This finding indicates that parameter adaptation allows OKAEM to continuously improve its optimization performance as new knowledge accumulates.}
\end{itemize}

These results collectively demonstrate OKAEM's robustness and adaptability across diverse knowledge transfer scenarios, validating its effectiveness in leveraging prior knowledge for enhanced EAs. \textcolor{black}{In addition, we present loss curves of pre-training and self-tuning, along with convergence curves of OKAEM and its variants (see Supplementary Appendix A), to further demonstrate the effectiveness of the proposed method.}

OKAEM exhibits the capability to learn from different types of prior knowledge. For each STOP, we generate diverse prior knowledge using GA \cite{2}, particle swarm optimization (PSO) \cite{488968}, and CMAES \cite{18} on source tasks, respectively. Detailed parameter settings are provided in Supplementary Appendix A. Fig. \ref{fig2}b illustrates the performance comparison between OKAEM, trained with different types of prior knowledge, and the corresponding source algorithms on target tasks. In all cases, OKAEM outperforms the source algorithms, indicating its capability to leverage various types of prior knowledge to enhance optimization efficiency. This robust performance gain highlights OKAEM's adaptability and versatility in enhancing optimization outcomes using different strategies. 

\textcolor{black}{In mixed-similarity scenarios (e.g., STOP6), source tasks exhibit heterogeneous similarity distributions $ h_2^m $ (linearly increasing similarity, see Supplementary Appendix A). This creates conflicting patterns in the prior knowledge: while some source tasks provide beneficial knowledge for transfer, others introduce divergent evolutionary behaviors that hinder convergence. In STOP6 (see Fig.2b), the suboptimal performance of OKAEM-CMAES compared to CMAES may stem from the failure of OKAEM's pre-training phase to capture the complex underlying patterns of CMAES-generated optimization knowledge across mixed-similarity sources.}

\begin{figure*}[htbp]
\centering
\includegraphics[width=0.82\textwidth]{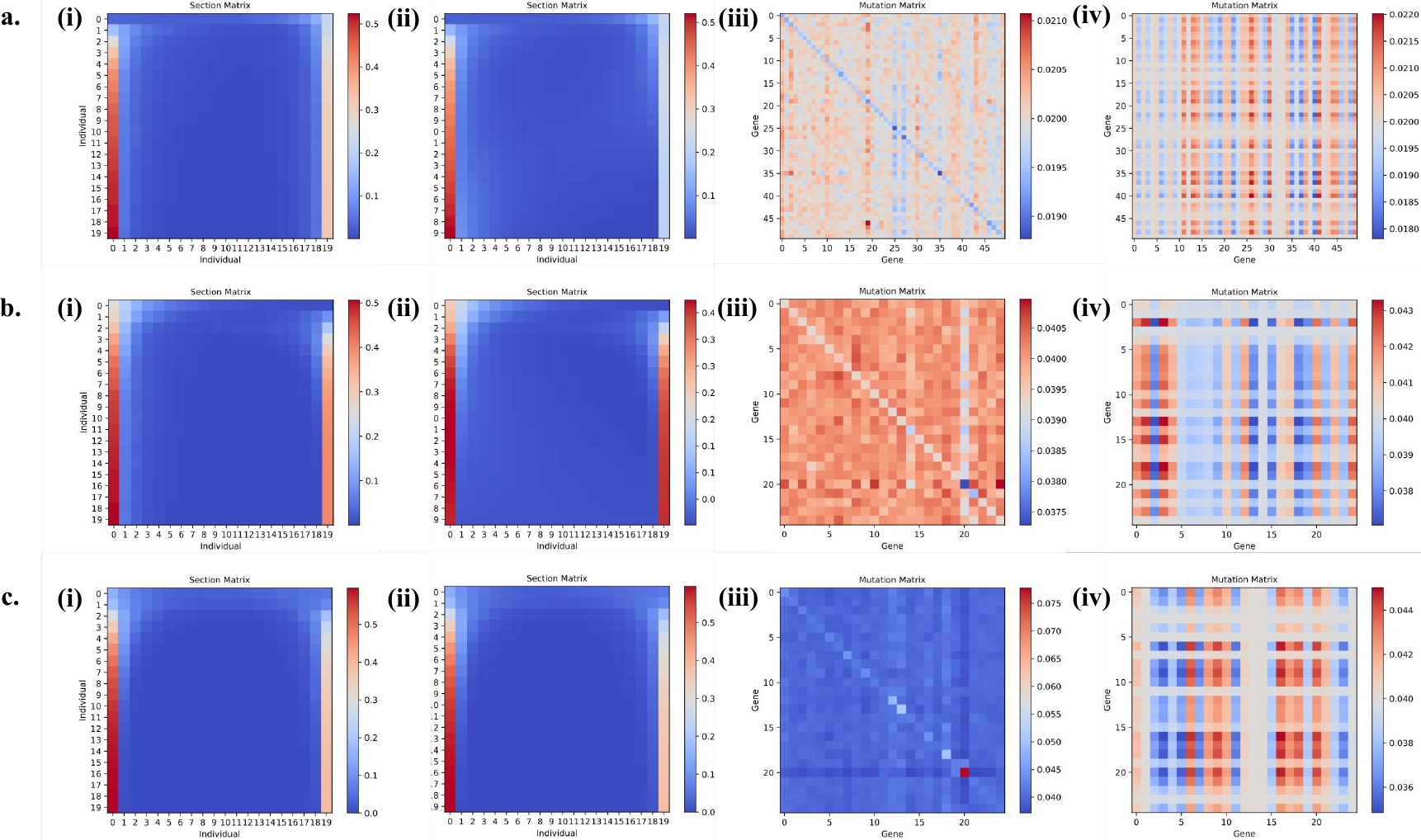}
\caption{Visualization of selection and mutation matrices learned by OKAEM on STOP1, STOP5, and STOP9.
For the selection matrix, axes represent individuals ranked by their fitness scores, with 0 and 19 denoting the best and worst individuals, respectively. The matrix values are computed as the mean across all $H$ attention heads.
For the mutation matrix, axes represent gene (decision variable) indices. a. STOP1: (i)-(ii) First and \textit{T}-th generation selection matrices; (iii)-(iv) First and \textit{T}-th generation mutation matrices.
b. STOP5: (i)-(ii) First and \textit{T}-th generation selection matrices; (iii)-(iv) First and \textit{T}-th generation mutation matrices.
c. STOP9: (i)-(ii) First and \textit{T}-th generation selection matrices; (iii)-(iv) First and \textit{T}-th generation mutation matrices.}\label{fig3}
\end{figure*}

We visualize the selection and mutation matrices learned by OKAEM for STOP1, STOP5, and STOP9 (Fig. \ref{fig3}). Across all cases, these matrices exhibit clear statistical patterns. For the selection matrix, individuals with higher fitness values are preferentially selected, embodying the principle of ``survival of the fittest''. Notably, individuals with the lowest fitness are selected more frequently than those with mid-range fitness, likely to maintain population diversity. The mutation matrix evolves from random to ordered over generations, exhibiting ``row-similarity'' characteristics, which indicate consistent patterns in gene variation. This evolution enhances the interpretability of our method compared to existing LEAs. In addition, ablation studies (see Table V in Supplementary Appendix A) show that eliminating either the crossover or mutation mechanisms results in a substantial performance drop, highlighting their indispensable roles in driving OKAEM's superior performance.

\subsection{Black-box optimization benchmarking problem}

\begin{figure*}[htbp]
\centering
\includegraphics[width=0.9\textwidth]{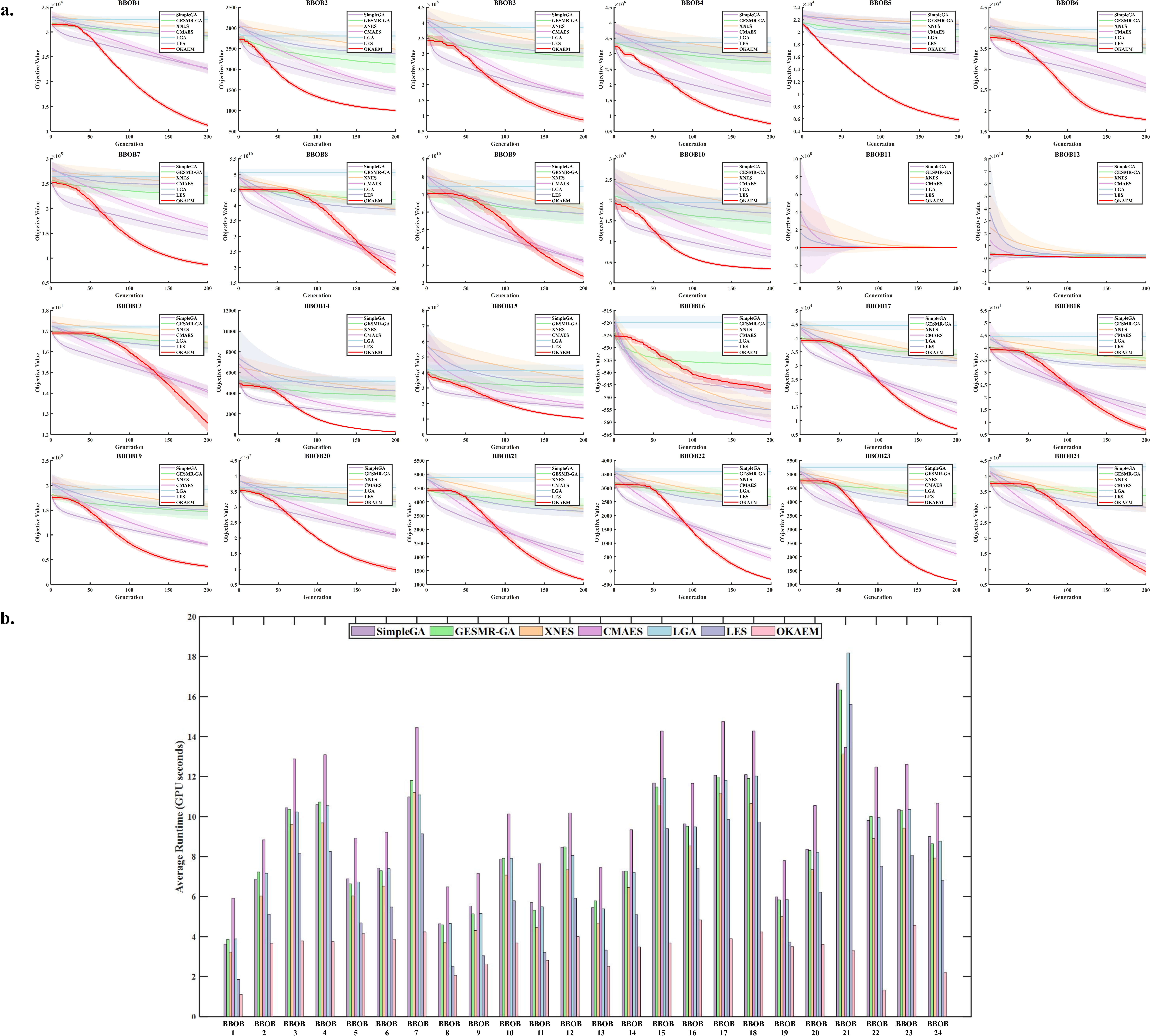}
\caption{Experimental results on the BBOB suite (20 independent runs with random initialization, population size $N=20$, number of iterations $T=200$). a. Average convergence curves. b. Average GPU runtime. \textcolor{black}{We also show loss curves of OKAEM’s adaptive optimization phase on the BBOB suite, which illustrates the stability of its self-tuning mechanism. The experimental results are shown in Supplementary Appendix B.}}\label{fig4}
\end{figure*}

\subsubsection{Problem configuration}

In many practical scenarios, collecting prior knowledge can be challenging, rendering pre-training infeasible. Leveraging its flexibility, OKAEM can perform adaptive optimization independently to solve complex tasks. This section evaluates OKAEM's performance without pre-training on the black-box optimization benchmarking (BBOB) suite \cite{Hansen02012021,10.1145/1830761.1830790}, where the search space is $[-10, 10]^d$ with $d = 1000$. The BBOB suite encompasses a series of high-dimensional continuous optimization functions, including unimodal, multimodal, rotated, and shifted functions, as well as those with specific properties such as Lipschitz continuity and second-order differentiability \footnote{The BBOB implementation is provided at \url{https://numbbo.github.io/coco/testsuites/bbob}.}. 

\subsubsection{Baseline}

This paper aims to advance the development of population-based EAs. Therefore, we do not compare with non-population-based approaches such as Bayesian optimization, which struggles with continuous optimization problems exceeding 100 dimensions. Furthermore, we omit large language models-based approaches \cite{10.1145/3694791,Lehman2024,yang2024large} from our baseline, as they are better suited for specific types of tasks. Our baseline includes classical EAs and their adaptive variants, as well as advanced LEAs.

\textbf{Classical EAs and their adaptive variants:}
\begin{itemize}
    \item SimpleGA \cite{such2017deep}, one of the most popular EAs in black-box optimization.
    \item GA with group elite selection of mutation rates (GESMR-GA) \cite{10.1145/3512290.3528706}, an advanced adaptive variant of GA.
    \item CMAES \cite{18} and XNES \cite{10.5555/2627435.2638566}, widely regarded as state-of-the-art EAs for addressing challenging continuous optimization problems (e.g., ill-conditioned, non-convex, non-continuous, or multi-modal).
\end{itemize}

\textbf{LEAs:}
\begin{itemize}
    \item Learnable GA (LGA) \cite{19}, which discovers new GAs in a data-driven manner, applicable to unseen optimization problems, search dimensions, and evaluation budgets.
    \item Learnable ES (LES) \cite{20}, which uses data-driven approaches to discover novel ESs with strong generalization and search efficiency.
\end{itemize}

\subsubsection{Parameter setting}

All experiments are conducted on a Linux platform with a GPU 2080Ti (Memory: 12 GB, CUDA version: 11.3). The implementations of baselines are obtained from a JAX-based evolutionary strategies library \footnote{The Python code for all comparison baselines is provided at \url{https://github.com/RobertTLange/evosax}.} \cite{10.1145/3583133.3590733}. For SimpleGA, GESMR-GA, CMAES, and XNES, the primary control parameters are automatically tuned. Other hyperparameters are optimized using grid search to identify the best combinations. For LGA and LES, we used the pre-trained parameters provided by the authors. All baselines are configured with a population size of 20 and a maximum of 200 generations. Detailed parameter settings are listed in Table \ref{tab:parameters}. \textcolor{black}{The parameter settings for OKAEM can be found in Supplementary Appendix B.}

\begin{table*}[htbp]
\centering
\caption{Detailed parameter configurations for the baselines. $ub$ and $lb$ are the upper and lower bounds of the problem, respectively. $\text{randn}(d)$ stands for sampling a $d$-dimensional vector from a standard normal distribution.}
\label{tab:parameters}
\footnotesize
\begin{tabular}{ccc}
\toprule
\textbf{Algorithm} & \textbf{Parameter} & \textbf{Setting} \\ \midrule
SimpleGA          & Initial $\sigma=0.2$ & \begin{tabular}[t]{@{}l@{}}We use grid search in $[0.1, 1]$ with \\ a step size of 0.1.\end{tabular} \\ 
                  & Crossover probability $p_c=0.7$ & \begin{tabular}[t]{@{}l@{}}We use grid search in $[0.5, 1]$ with \\ a step size of 0.1.\end{tabular} \\ \midrule
GESMR-GA         & Initial $\sigma=0.2$ & \begin{tabular}[t]{@{}l@{}}We use grid search in $[0.1, 1]$ with \\ a step size of 0.1.\end{tabular} \\ \midrule
CMAES/XNES        & Initial $\sigma=0.2$ & \begin{tabular}[t]{@{}l@{}}We use grid search in $[0.1, 1]$ with \\ a step size of 0.1.\end{tabular} \\ 
                  & Initial $\mu$ & 
                  \begin{tabular}[t]{@{}l@{}}
                  $\mu = lb + \text{randn}(d) \times (ub - lb)$
                  \end{tabular} \\ \midrule
LGA               & All parameters & \begin{tabular}[t]{@{}l@{}}We use the pre-trained parameters \\ provided by the authors.\end{tabular} \\ \midrule
LES               & All parameters & \begin{tabular}[t]{@{}l@{}}We use the pre-trained parameters \\ provided by the authors.\end{tabular} \\ \bottomrule
\end{tabular}
\end{table*}

\subsubsection{Result}
\textcolor{black}{As observed in Fig. \ref{fig4}a, OKAEM achieves superior performance across most benchmarks, regardless of their diverse geometric properties. This indicates that even without prior knowledge (i.e., pre-training), OKAEM's adaptive optimization can deliver highly competitive results.} Unlike classical adaptive algorithms such as GESMR, XNES, and CMAES, which rely on specific structures of evolutionary operators, OKAEM's parameter adaptation is entirely driven by population dynamics and fitness. This flexibility makes it suitable for any neural representation-based evolutionary operator. By adjusting its parameters based on population dynamics, OKAEM outperforms all baselines, demonstrating strong adaptability and optimization capabilities. Conversely, LGA and LES rely on fixed parameters post-training, preventing them from adapting to population changes. \textcolor{black}{As demonstrated by OKAEM's outstanding results, adaptability is crucial for enhancing the performance of EAs.} Moreover, Fig. \ref{fig4}b shows that OKAEM has the lowest computational cost among all baselines, requiring only a few GPU seconds. This indicates that neural representation-based OKAEM can significantly enhance computational efficiency through parallel computing.

\subsection{Black-box prompt tuning for the vision-language model}

\begin{figure*}[htbp]
\centering
\includegraphics[width=0.75\textwidth]{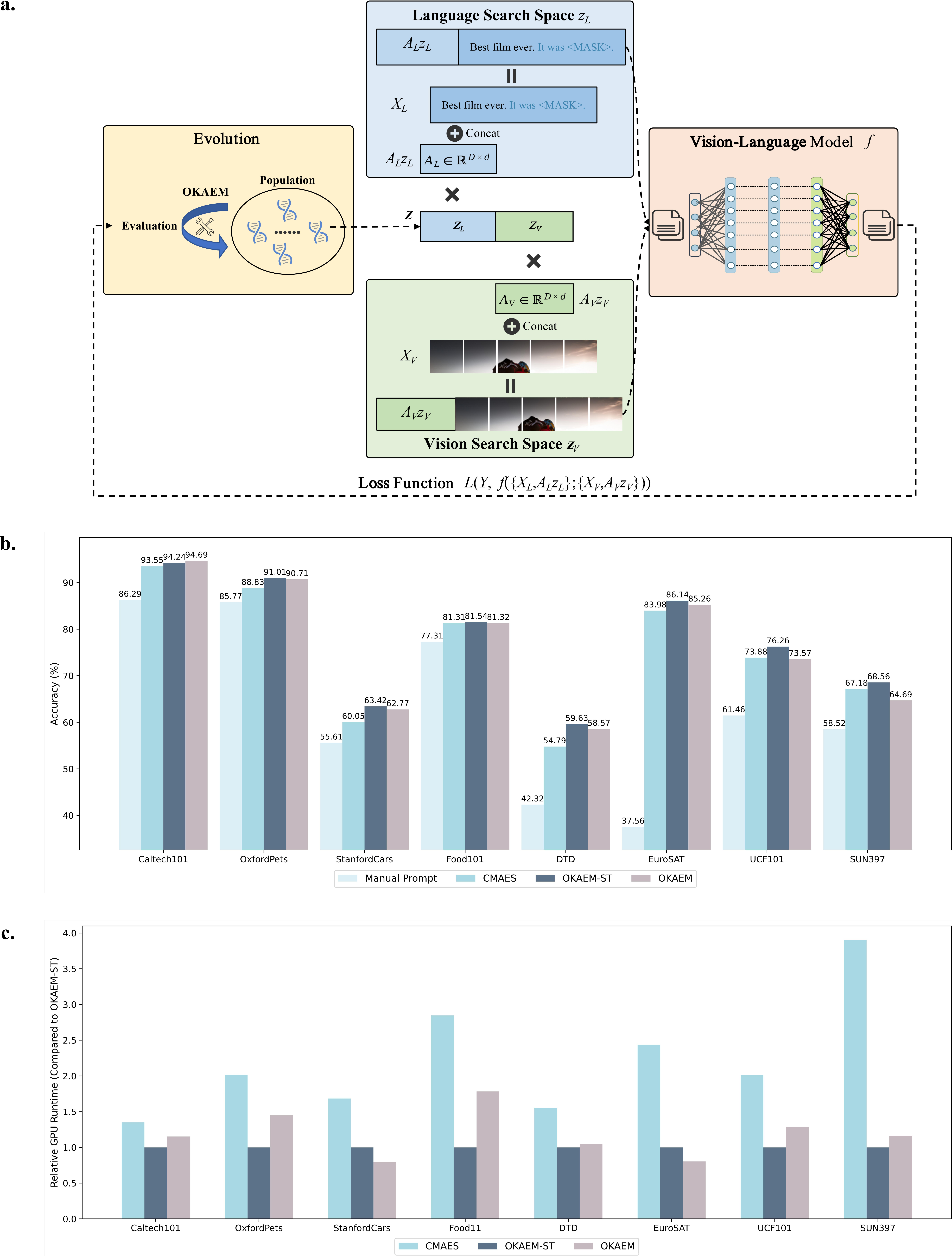}
\caption{Black-box prompt tuning for vision-language models. a. Illustration of the prompt tuning process. b. Average test accuracies on 8 common visual image classification tasks over five independent runs. c. Relative GPU runtime of the automatic tuning methods compared to OKAEM-ST.}\label{fig5}
\end{figure*}

\subsubsection{Problem}
Pre-trained models, particularly those for vision-language tasks, are commonly released as services that allow users to set task-specific prompts to query the models \cite{ijcai2023p187}. EAs such as CMAES are widely used to optimize these prompts to enhance the performance of pre-trained models \cite{pmlr-v162-sun22e,sun-etal-2022-bbtv2,ijcai2023p187,chao2024match}. Without access to the model's architecture or gradient information, optimizing prompts in this black-box setting remains a significant challenge. Let the forward propagation of a vision-language model be denoted as $ f $. Given batches of texts $ X_L $ and images $ X_V $, along with respective text prompts $ p_L $ and image prompts $ p_V $, the model outputs a similarity score for each image-text pair. As shown in Fig.~\ref{fig5}a, with outputs and labels $ Y $, the objective of prompt tuning is to minimize the cross-entropy loss $ L $:
\begin{equation}
    Z^* = \arg \min_{Z = \{z_L, z_V\}} L(Y, f(\{X_L, p_L\}; \{X_V, p_V\})),
\end{equation}
where
$$
p_L = A_L Z_L, \quad p_V = A_V Z_V,
$$
and $Z$ represents the parameter subspace for prompts to be optimized, composed of the text intrinsic vector $Z_L$ and visual intrinsic vector $Z_V$. The randomly initialized fixed matrices $ A_L $ and $ A_V $ project $ Z_L $ and $ Z_V $ into the text prompt $ p_L $ and image prompt $ p_V $, respectively. This optimization objective is computed using only the forward pass of the vision-language model, eliminating the need for backpropagation.

\textcolor{black}{While OKAEM is a general-purpose optimizer applicable to any black-box problem, we select black-box prompt tuning for vision-language models as a representative real-world application for three reasons:}  

\begin{itemize}
    \item \textcolor{black}{Vision-language models represent a frontier research area in machine learning with growing industrial applications (e.g., zero-shot classification, cross-modal retrieval);}  
    \item \textcolor{black}{Prompt tuning provides a clean black-box optimization interface through forward-pass-only evaluation;}  
    \item \textcolor{black}{The high-dimensional parameter space (over 1000 dimensions) and complex loss landscapes pose significant challenges to the adaptability of traditional EAs.}
\end{itemize}

\subsubsection{Dataset}

\begin{table*}[htbp]
\centering
\caption{Statistics of the datasets for prompt tuning in vision-language models.}
\label{tab:dataset-stats}
\footnotesize
\begin{tabular}{ccccccc}
\toprule
\textbf{Dataset} & \textbf{Classes} & \textbf{Train} & \textbf{Val} & \textbf{Test} & \textbf{Hand-crafted prompt} & \begin{tabular}[t]{@{}l@{}}\textbf{Maximum number} \\ \textbf{of evaluations}\end{tabular} \\
\midrule
Caltech101 & 100 & 4,128 & 1,649 & 2,465 & ``a photo of a [CLASS].'' & 4,800 \\
OxfordPets & 37 & 2,944 & 736 & 3,669 & ``a photo of a [CLASS], a type of pet.'' & 4,800 \\
StanfordCars & 196 & 6,509 & 1,635 & 8,041 & ``a photo of a [CLASS].'' & 25,200 \\
Food101 & 101 & 50,500 & 20,200 & 30,300 & ``a photo of [CLASS], a type of food.'' & 4,800 \\
UCF101 & 101 & 7,639 & 1,898 & 3,783 & ``a photo of a person doing [CLASS].'' & 25,200 \\
SUN397 & 397 & 15,880 & 3,970 & 19,850 & ``a photo of a [CLASS].'' & 18,000 \\
EuroSAT & 10 & 13,500 & 5,400 & 8,100 & ``a centered satellite photo of [CLASS].'' & 25,200 \\
DTD & 47 & 2,820 & 1,128 & 1,692 & ``[CLASS] texture.'' & 36,000 \\
\bottomrule
\end{tabular}
\end{table*}

To validate OKAEM's capability to solve complex real-world problems, we evaluate its performance on eight commonly used visual image classification datasets: Caltech101 \cite{fei2004learning}, OxfordPets \cite{parkhi2012cats}, StanfordCars \cite{Krause_2013_ICCV_Workshops}, Food101 \cite{10.1007/978-3-319-10599-4_29}, UCF101 \cite{soomro2012ucf101}, SUN397 \cite{5539970}, EuroSAT \cite{8736785}, and DTD \cite{6909856}. These datasets, covering a wide range of visual tasks, are commonly used to evaluate prompt tuning tasks. Their statistics are summarized in Table \ref{tab:dataset-stats}.

\begin{itemize}
    \item \textbf{Caltech101}, \textbf{OxfordPets}, and \textbf{StanfordCars} provide images for fine-grained classification and recognition. Caltech101 includes images of 101 object categories; OxfordPets features a diverse set of pet breeds; and StanfordCars offers detailed images of car models for precise vehicle identification.
    
    \item \textbf{Food101} focuses on food image classification, covering a broad spectrum of global cuisines.
    
    \item \textbf{UCF101} specializes in human action recognition from video clips, assessing both category-specific and dynamic scene understanding.
    
    \item \textbf{SUN397} provides a nearly exhaustive collection of scenes, suitable for evaluating scene recognition tasks.

    \item \textbf{EuroSAT} comprises satellite imagery for land use classification, representing various Earth surface covers from a spaceborne perspective.
    
    \item \textbf{DTD} targets the visual recognition of diverse textural patterns found in both natural and artificial surfaces.
    
\end{itemize}

\subsubsection{Baseline}
We compare OKAEM with three types of methods: 1) Manual prompt \cite{pmlr-v139-radford21a}, which uses carefully crafted templates for zero-shot evaluation. 2) CMAES \cite{ijcai2023p187,chao2024match}, a state-of-the-art black-box prompt tuning algorithm widely used to optimize prompts. 3) OKAEM-ST, a variant of OKAEM that performs only adaptive optimization without pre-training. Additionally, we employ CMAES for prompt tuning on the Caltech101 dataset to generate prior knowledge for pre-training OKAEM. Once trained, OKAEM directly tunes prompts across all datasets without separate pre-training for each one. 

\subsubsection{Problem configuration}

In alignment with the experimental setup in \cite{zhou2022learning}, the same 16-shot split is used for prompt tuning in all methods. Evaluation is carried out on full test sets for comparison.

Following the configuration of the model adopted in \cite{ijcai2023p187}, the open-source CLIP model with ViT-B/32 is used as the backbone for the visual encoder. The intrinsic dimension is set to 1000. The vision prompt length and language prompt length are set to 8 and 5, respectively.

\subsubsection{Parameter setting}

All experiments are conducted on a Linux platform with a GPU 2080Ti (Memory: 12 GB, CUDA version: 11.3) \footnote{The prompt tuning for visual-language models is detailed in \url{https://github.com/BruthYU/BPT-VLM}.}. The general settings are as follows:
\begin{itemize}
    \item Loss function: Cross entropy.
    \item Number of independent runs: 5.
    \item Population size $N$: 12 for SUN397 to prevent memory overflow during parallel computing, and 20 for other datasets.
    \item Interval for calculating test accuracy: Every 12 evaluations.
    \item Maximum number of evaluations $F$: See Table \ref{tab:dataset-stats}.
\end{itemize}

The configuration of the manual prompt is consistent with \cite{pmlr-v139-radford21a}. The main control parameters of CMAES are automatically adjusted. Other hyperparameters of CMAES are tuned using grid search to determine the optimal combination (see Table \ref{tab:parameters} for specific settings).
\textcolor{black}{The parameter settings for all methods can be found in Supplementary Appendix C.}

\subsubsection{Result}
Fig. \ref{fig5}b illustrates that black-box prompt tuning methods consistently outperform the manual prompt, underscoring the importance and efficacy of automated tuning. After multiple function evaluations, OKAEM-ST achieves higher accuracy on most datasets compared to the baselines. Notably, OKAEM demonstrates an advantage over OKAEM-ST on the Caltech101 dataset. This is attributed to the prior knowledge derived from Caltech101, highlighting the improvement in model performance when the prior knowledge closely aligns with the target task. Moreover, Fig. \ref{fig5}c shows that both OKAEM and OKAEM-ST exhibit lower computational costs compared to CMAES, significantly accelerating evolutionary computation in practical applications.

\subsection{Parameter sensitivity analysis}

\begin{figure*}[htbp]
\centering
\includegraphics[width=0.8\textwidth]{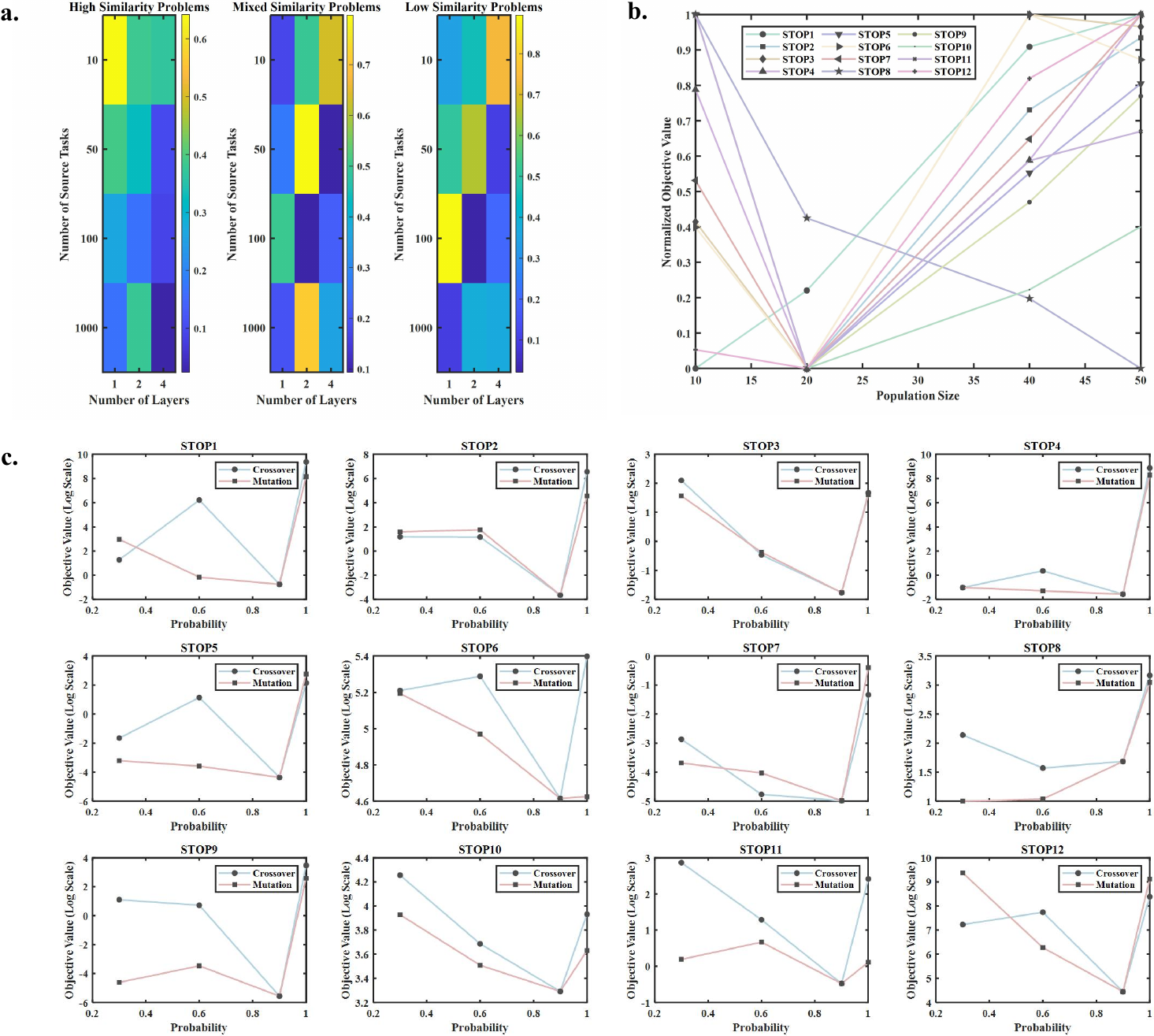}
\caption{Sensitivity analysis of key parameters in OKAEM (20 independent runs with random initialization). The stopping
criterion is set to a maximum of 5000 evaluations. a. Analysis of layers $L\in [1,2,4]$ and number of source tasks $K\in [10,50,100,1000]$. The heatmap displays the average normalized objective values across different types of STOPs, with all values scaled to the $[0, 1]$ using min-max normalization. b.  Population size $N\in[10, 20, 40, 50]$. The y-axis shows the normalized objective values. c. Dropout probability 
$p_C\in[0.3, 0.6, 0.9, 1]$ in crossover and dropout probability 
$p_M\in[0.3, 0.6, 0.9, 1]$ in mutation. The x-axis represents the objective values on a log scale.}\label{fig6}
\end{figure*}

Fig.~\ref{fig6} presents the sensitivity analysis of OKAEM's four critical parameters on the STOP suite: the number of layers $L$, population size $N$, dropout probability $p_C$ in crossover, and dropout probability $p_M$ in mutation. Fig.~\ref{fig6}a illustrates that the performance of OKAEM improves with an increase in both the number of source tasks and model depth, regardless of whether the task similarity is high (STOP1-STOP4) or low (STOP9-STOP12). A detailed experimental setup is provided in Supplementary Appendix A.

\textcolor{black}{More source tasks provide richer prior knowledge, allowing deeper models to capture complex patterns. The positive correlation between performance and prior knowledge is especially pronounced in highly similar contexts (STOP1-STOP4). This implies that OKAEM's optimization capacity strengthens with the accumulation of prior knowledge. Conversely, in mixed-similarity conditions (STOP5-STOP8), the relationship between source tasks, model depth, and performance becomes more complex. For these cases, we recommend manually tuning the model depth for optimal results.}

As indicated in Fig.~\ref{fig6}b, a population size of 20 is recommended when the maximum number of evaluations is 5000. In practice, the population size should be adjusted based on the problem dimensions. Fig.~\ref{fig6}c illustrates that initial increases in $ p_C $ and $ p_M $ improve the performance of OKAEM. However, performance degrades when these parameters exceed critical thresholds. Specifically, setting $ p_C $ or $ p_M $ to 1 eliminates randomness in crossover or mutation, leading to suboptimal performance. \textcolor{black}{This confirms the critical role of stochasticity in maintaining search diversity and convergence stability.} Based on experimental results, we recommend using moderately high values for $ p_C $ and $ p_M $ to balance performance optimization and stability.

\section{Conclusion}\label{sec13}

With the advancement in computational capabilities, complex optimization tasks in scientific and industrial fields have become increasingly intricate and challenging. Traditional EAs often rely heavily on specific problem structures, limiting their ability to leverage the vast amount of valuable knowledge generated during optimization. This constrained transferability and adaptability hinder their optimization performance and reduce confidence in practical applications. This paper introduces a novel neural representation-based evolutionary framework, OKAEM, which efficiently utilizes prior knowledge and quickly adapts to self-generated real-time knowledge. 

As demonstrated across the STOP suite, the BBOB suite, and a real-world case study, OKAEM exhibits superior optimization performance compared to existing baselines, owing to its robust transferability and adaptability. Extensive experiments show that OKAEM exhibits strong learnability: 1) Its optimization capability improves as knowledge accumulates; 2) It can explicitly learn the principles of natural selection and genetic combination in evolution. Thus, OKAEM takes a significant step forward in modeling transferability, adaptability, and learnability in EAs, overcoming the inflexibility of existing customized methods and providing a foundational model for addressing larger-scale complex optimization tasks. \textcolor{black}{For more case studies and practical usage tips, please refer to Supplementary Appendices F and G.} Despite the progress made by OKAEM, several avenues for improvement remain: 

\begin{itemize}
    \item Experimental results (Fig. \ref{fig2}a) show that in certain mixed-similarity scenarios (STOP6 and STOP8), OKAEM outperforms methods without knowledge transfer but falls short of those specifically designed for "what-to-transfer". This highlights the potential benefits of refining pre-training datasets to better match target tasks. 

    \item Integrating end-to-end training with fitness evaluation environments could enhance OKAEM's performance in real-world applications by leveraging the synergy between evolution and evaluation. 

    \item \textcolor{black}{Coupling ESTO and MetaBBO by leveraging prior knowledge modeling to design training tasks presents a promising direction to enhance MetaBBO generalization.}
    
    \item The current framework is not tailored for multi-objective optimization, a common application area for EAs \cite{996017,4358754}. Constructing prior knowledge, model architectures, and training paradigms for multi-objective optimization merits further exploration. 
    
    \item Selecting foundational source optimization tasks when constructing prior knowledge datasets can significantly improve model generalization, potentially leading to large-scale pre-trained models capable of addressing diverse optimization challenges. 
    
    \item \textcolor{black}{A rigorous convergence rate analysis of OKAEM on certain problems (such as quadratic functions) is crucial for establishing theoretical guarantees.} 
    
    \item \textcolor{black}{Current experiments focus on 1000-dimensional problems from the BBOB suite and black-box prompt tuning tasks for vision-language models, establishing feasibility in high-dimensional settings. For extreme-scale problems ($d>10^4$), attention mechanisms with reduced computational complexity (e.g., linearized variants) are expected to enhance scalability. \textcolor{black}{In addition, exploring different mechanisms for implementing randomness in neural evolutionary operators is a promising direction.}} 

\end{itemize}

It should be noted that these challenges present engineering opportunities rather than insurmountable barriers.

\bibliographystyle{IEEEtran}
\bibliography{my}

\begin{IEEEbiography}[{\includegraphics[width=1in,height=1.25in,clip,keepaspectratio]{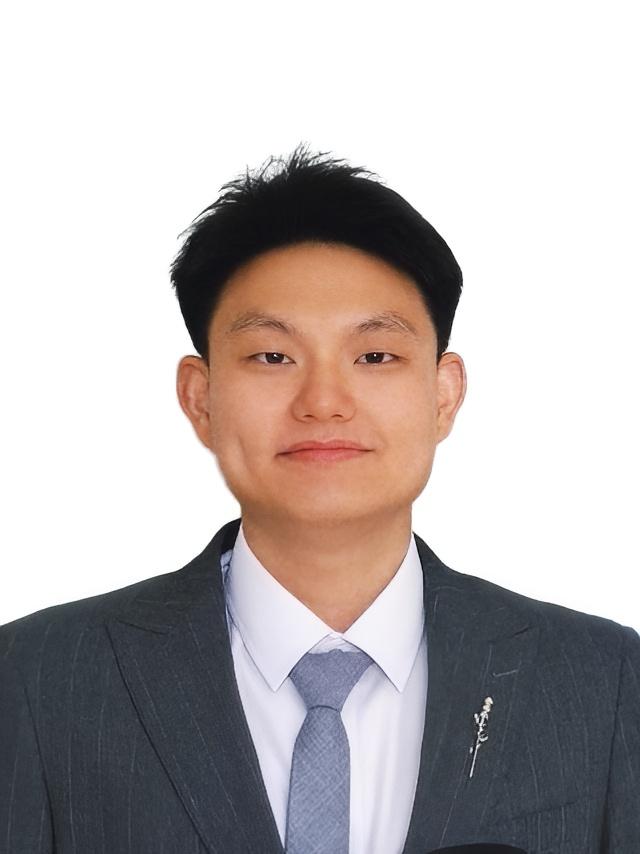}}]{Chao Wang} received the B.S. degree in intelligent science and technology and the Ph.D. degree in computer science and technology from Xidian University, Xi’an, China, in 2019 and 2025, respectively. He is currently a Post-Doctoral Researcher with the Key Laboratory of Intelligent Perception and Image Understanding, Ministry of Education,
Xidian University. His research interests include
genetic and evolutionary computing, evolutionary machine learning, and black-box optimization.

\end{IEEEbiography}

\begin{IEEEbiography}[{\includegraphics[width=1in,height=1.25in,clip,keepaspectratio]{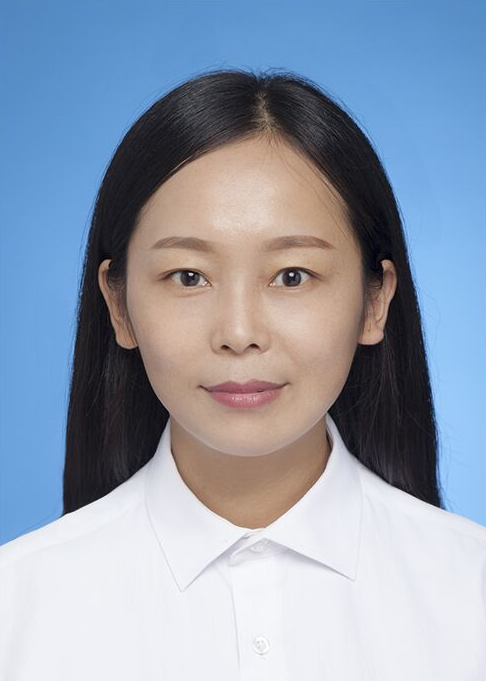}}]{Lingling Li}
(Senior Member, IEEE) received the B.S. and Ph.D. degrees from Xidian University, Xi’an, China, in 2011 and 2017, respectively.

From 2013 to 2014, she was an exchange Ph.D. student with the Intelligent Systems Group, Department of Computer Science and Artificial Intelligence, University of the Basque Country UPV/EHU, Spain. She is an associate Professor with the Key Laboratory of Intelligent Perception and Image Understanding, Ministry of Education, School of Artificial Intelligence, Xidian University. Her research interests include image processing, deep learning, and pattern recognition.
\end{IEEEbiography}
\vspace{-1cm}

\begin{IEEEbiography}[{\includegraphics[width=1in,height=1.25in,clip,keepaspectratio]{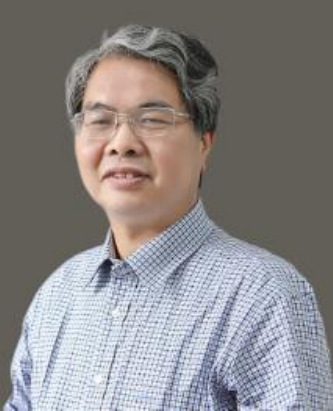}}]
{Licheng Jiao} (Fellow, IEEE) received his B.S. degree from Shanghai Jiaotong University, Shanghai, China, in 1982, and his M.S. and PhD degrees from Xi’an Jiaotong University, Xi’an, China, in 1984 and 1990, respectively.

Since 1992, he has been a professor at Xidian University. Now, he is a distinguished professor and serves in the School of Artificial Intelligence, Xidian University, Xi’an. He currently serves as the Director of the Key Laboratory of Intelligent Perception and Image Understanding, which is affiliated with the Ministry of Education of China. He has been a member of the Academia Europaea. His research interests include machine learning, deep learning, natural computation, remote sensing, image processing, and intelligent information processing.
\end{IEEEbiography}
\vspace{-1cm}

\begin{IEEEbiography}[{\includegraphics[width=1in,height=1.25in,clip,keepaspectratio]{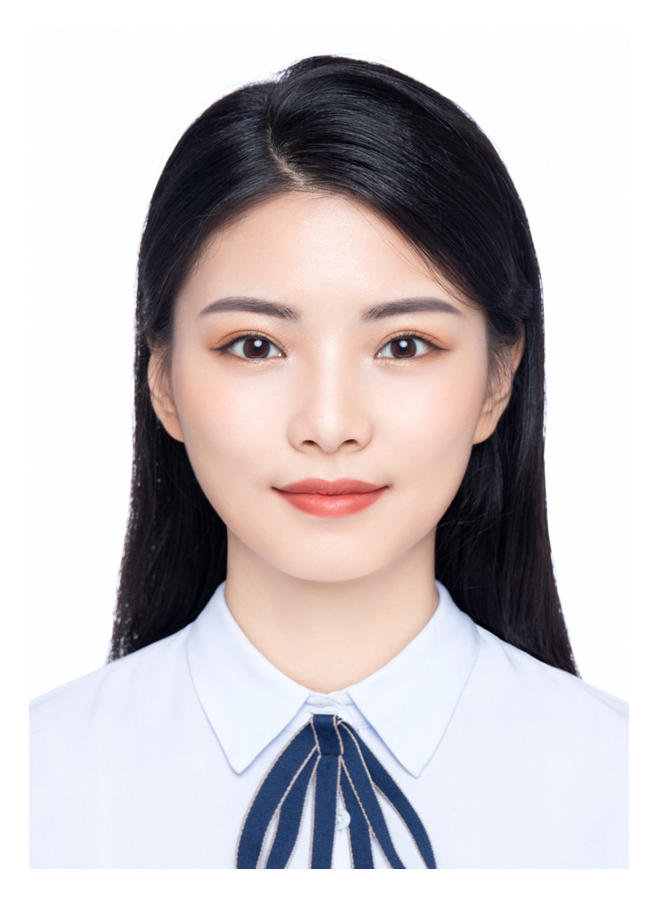}}]{Jiaxuan Zhao} (Student Member, IEEE) received the B.S. degree in materials science and engineering from Xidian University, Xi’an, China, in 2019. She is currently pursuing the Ph.D. degree with the Key Laboratory of Intelligent Perception and Image Understanding of Ministry of Education, School of Artificial Intelligence Xidian University, Xi’an, China. 

Her research interests include multimodal fusion, evolutionary computing, and image understanding.
\end{IEEEbiography}

\begin{IEEEbiography}[{\includegraphics[width=1in,height=1.25in,clip,keepaspectratio]{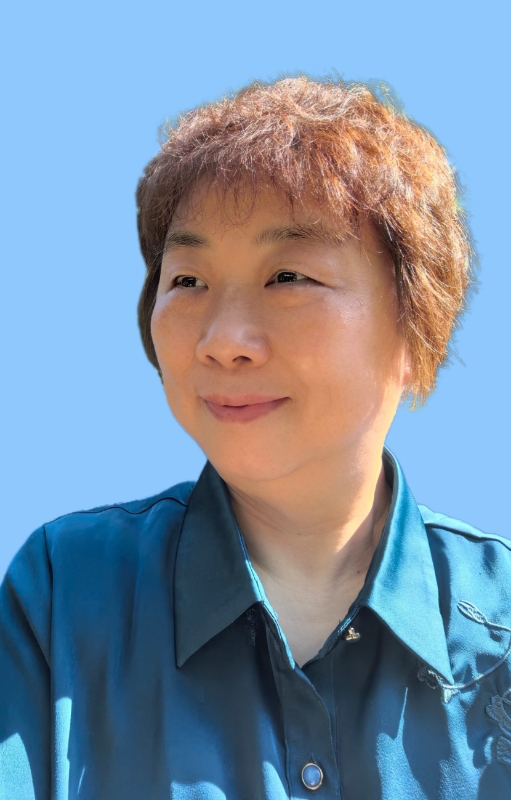}}]{Fang Liu}
(Senior Member, IEEE) received the B.S. degree in computer science and technology from the Xi’an Jiaotong University, Xi’an, China, in 1984 and the M.S. degree in computer science and technology from the Xidian University, Xi’an, in 1995. 

She is currently a Professor with the School
of Artificial Intelligence, Xidian University. Her
research interests include signal and image processing, synthetic aperture radar image processing,
multi-scale geometry analysis, learning theory and
algorithms, optimization problems, and data mining.
\end{IEEEbiography}
\vspace{-1cm}

\begin{IEEEbiography}[{\includegraphics[width=1in,height=1.25in,clip,keepaspectratio]{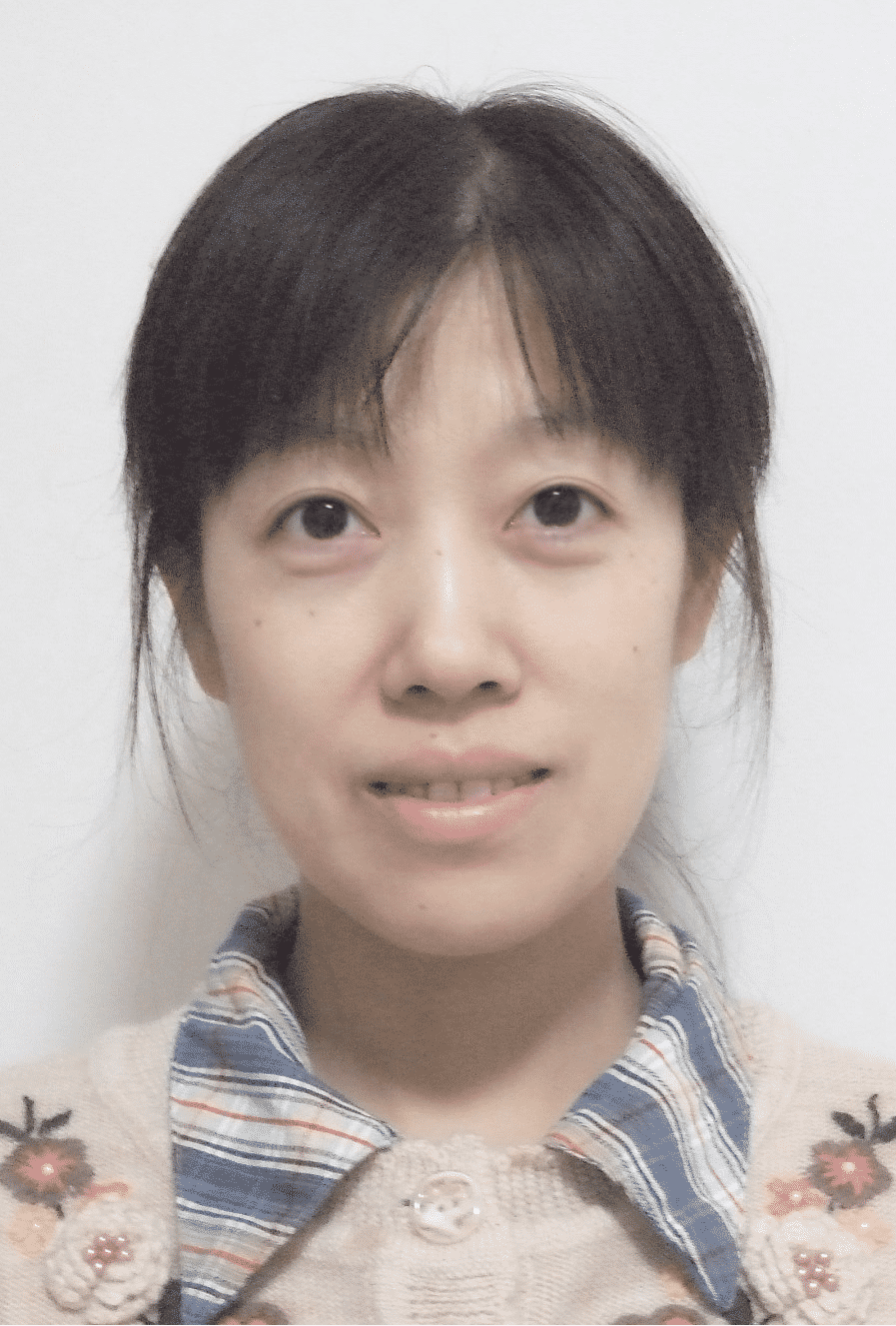}}]{Shuyuan Yang}
(Senior Member, IEEE) received the
B.A. degree in electrical engineering, and the M.S.
and Ph.D. degrees in circuit and system from Xidian
University, Xi’an, China, in 2000, 2003, and 2005,
respectively.

She has been a Professor with the School
of Artificial Intelligence, Xidian University. Her
research interests include machine learning and
multiscale geometric analysis.
\end{IEEEbiography}

\end{document}